\documentclass{article}

\usepackage{arxiv}

\usepackage{newtxtext}
\usepackage[utf8]{inputenc} 
\usepackage[T1]{fontenc}    
\usepackage{hyperref}       
\usepackage{url}            
\usepackage{booktabs}       
\usepackage{amsfonts}       
\usepackage{nicefrac}       
\usepackage{microtype}      
\usepackage{lipsum}         
\usepackage{graphicx}
\usepackage[numbers]{natbib}
\usepackage{doi}

\usepackage{amsmath}
\usepackage{algorithm}
\usepackage{algorithmic}
\usepackage[caption=false,font=normalsize,labelfont=sf,textfont=sf]{subfig}

\usepackage[capitalize]{cleveref}
\Crefname{section}{Section}{Sections}
\Crefname{figure}{Figure}{Figures}
\Crefname{table}{Table}{Tables}
\Crefname{definition}{Definition}{Definitions}
\Crefname{theorem}{Theorem}{Theorems}
\Crefname{corollary}{Corollary}{Corollaries}
\Crefname{algorithm}{Algorithm}{Algorithms}
\Crefname{equation}{Equation}{Equations}

\newtheorem{Definition}{Definition}

\newtheorem{Lemma}{Lemma}

\newtheorem{Corollary}{Corollary}

\newtheorem{Example}{Example}

\usepackage{amssymb}
\newcommand{\D}{\mathcal{D}}
\newcommand{\X}{\mathcal{X}}
\newcommand{\Y}{\mathcal{Y}}
\newcommand{\Z}{\mathcal{Z}}
\newcommand{\Hs}{\mathcal{H}}
\newcommand{\E}{\mathbb{E}}

\newcommand{\R}{\mathbb{R}}
\newcommand{\Nat}{\mathbb{N}}
\newcommand{\set}[1]{\{#1\}}
\newcommand{\loss}{\ell}
\renewcommand{\P}{\mathbb{P}}

\usepackage{xcolor}

\title{Monotonic Learning in the PAC Framework: A New Perspective}

\author{
Ming Li \\
Jinan University \\
\AND
Cheny Zhang \\
University of Canterbury \\
\texttt{chenyi.zhang@canterbury.ac.nz}
\AND
Qin Li \\
East China Normal University \\
}

\renewcommand{\shorttitle}{\textit{arXiv} Template}

\begin{document}
\maketitle

\begin{abstract}
Monotone learning describes learning processes in which expected performance consistently improves as the amount of training data increases. However, recent studies challenge this conventional wisdom, revealing significant gaps in the understanding of generalization in machine learning. Addressing these gaps is crucial for advancing the theoretical foundations of the field. In this work, we utilize Probably Approximately Correct (PAC) learning theory to construct a theoretical risk distribution that approximates 
a learning algorithm's actual performance. We rigorously prove that this theoretical distribution exhibits monotonicity as sample sizes increase. We identify two scenarios under which deterministic algorithms based on Empirical Risk Minimization (ERM) are monotone: (1) the hypothesis space is finite, or (2) the hypothesis space has finite VC-dimension. 
Experiments on three classical learning problems validate our findings by demonstrating that the monotonicity of the algorithms' generalization error is guaranteed, as its theoretical risk upper bound monotonically converges to the minimum generalization error.
\end{abstract}


\section{Introduction}\label{sec:introduction}

A foundational premise of machine learning is that learners improve their performance by learning from empirical data~\cite{shalev2014understanding,mohri2018foundations}.
It is widely believed that increasing the amount of training data should enhance a model's performance or, at the very least, not degrade it. 
This expectation is reflected in much of theory and practice in machine learning, 
as \citet{shalev2014understanding} discuss in their book, the performance of a model begins to increase once the training set size exceeds the VC dimension of the hypothesis space. 
However, recent studies have challenged this assumption, revealing non-monotonic behavior (where adding more data can lead to worse performance) in various tasks, including density estimation, classification, and regression~\cite{loog2019minimizers}. 
Two prominent examples of non-monotonicity are  \emph{peaking}, where the generalization error reaches a maximum at a certain sample size before improving with additional data~\cite{vallet1989linear}, and \emph{dipping}, which is characterized by the generalization error achieving a minimum at a specific sample size before starting to increase for larger sample sizes~\cite{loog2012dipping}. 
These counterintuitive findings highlight significant gaps in our understanding of generalization. 

To better understand these phenomena, it is essential to revisit the theoretical foundations of generalization in machine learning. 
The Probably Approximately Correct (PAC) learning framework~\cite{valiant84} provides a foundational formalism for analyzing and quantifying generalization behavior. 
In this framework, the generalization performance of algorithms guided by the Empirical Risk Minimization (ERM) principle is captured by the sum of approximation error and estimation error. 
The approximation error is determined by the choice of hypothesis space, while the estimation error, influenced by the training set size, decreases as the training set grows larger~\cite{shalev2014understanding}. Although these theoretical bounds offer valuable insights, their practical application is often challenging, particularly when empirical results deviate from standard expectations. 
The existence of such cases
has led to critical examination of the assumptions underlying generalization bounds. Notably, \citet{viering2019open} pose a pivotal question that forms the focus of this paper: “What conditions must a learner satisfy to be monotone?”

In this work, we provide an alternative view on the learning process in the framework of PAC learning. 
Given a machine learning problem that is PAC learnable, the generalization error of the output model of the learning algorithm must be bounded by an arbitrarily small value $\epsilon > 0$ with a confidence $1-\delta$ for arbitrary small $\delta\in[0,1]$, provided that the size of a training set is sufficiently large. Fixing the size of the training set, an output model is more likely to exhibit a potentially larger generalization error as the probability increases (i.e., the model is more likely to meet the objective if the performance threshold is lowered). Analyzing the derivative of the mapping from $\epsilon$ to $1-\delta$ reveals a probability density function, or a corresponding distribution, that represents the upper bound on the generalization error of a learned model. Increasing the sample size effectively moves the distribution closer to the minimum generalization error (shown in \cref{fig:probability-2}). 

The risk bound provided by PAC is loose when a sample size is small. 
Consequently, it often serves as an upper bound on the actual generalization risk, or equivalently, a lower bound on the model's performance. 
In Section~\ref{sec:experiment} we conduct experiment on three  
learning tasks as examples, and compare the empirical distributions for the models trained in the experiment with the theoretical 
distributions predicted by PAC. We are able to show that the average performance values derived from the distributions are shown as monotone when depicted in 
learning curves. Moreover, the two distributions start to merge as the training sample sizes are sufficiently large. 
As expected, \cref{fig:CBL_QA}~(a) illustrates the learning curves derived from the means of the empirical distribution ($P_m$) and the theoretical distribution ($Q_m$), both of which demonstrate errors that decrease monotonically with high probability. 
Our approach extends the analysis from a single scalar bound analysis to a full distributional characterization and strengthens convergence guarantees from probabilistic limits to distribution-based monotonicity, thereby establishing a principled link between statistical learning theory and practical algorithms. 
Additionally, it enables systematic interpretability and robust stability analysis of learning algorithms under finite-sample conditions.

\paragraph{Layout of the paper.}
We introduce the basic concepts that are used in this paper in Section~\ref{sec:preliminaries}. Section~\ref{sec:pac-monotone} derives the lower bound risk distributions for two classes of PAC learnable problems. 
We then carry out experiments on three concrete PAC learnable tasks 
and compare experimental distributions with the PAC derived distributions for each problem in Section~\ref{sec:experiment}. 
Related works are reviewed in Section~\ref{sec:relatedwork}, and the paper is concluded in Section~\ref{sec:conclusion}.


\section{Preliminaries}\label{sec:preliminaries}

Let $S = \{ (x_1, y_1), (x_2, y_2), ... (x_m, y_m) \}$ be a data set of size $m$, sampled i.i.d. from 
a probability distribution $\D$ over $\X \times \mathcal{Y}$, where $\X$ is the input space (features) and $\mathcal{Y}$ is the output space (classification labels). 
A hypothesis space $\Hs\subseteq \Y^{\X}$ is always implicitly determined by a learning algorithm $A$. 
Given $h\in \Hs$ and $x\in\X$, $\hat{y} = h(x)$ is the prediction from the hypothesis $h$ on input $x$.
The performance of $h$ on a single input $x$ can be measured by using a loss function $\loss$ that compares $h(x)$ with $y$. 
Ideally, given $\loss$, the performance of a hypothesis $h$ is defined by \emph{generalization error} $L_{\D}(h)$ over the true distribution $\D$.
\begin{equation}
    L_{\D}(h) = \P_{(x,y) \sim \D}[h(x) \neq y] = \E_{(x,y) \sim \D} [\loss (h(x), y)]
\end{equation}

Since we focus on the classification problems in this paper, we have $L_{\D}(h) \in [0, 1]$ for any $h$.
Given it is often impractical to obtain the generalization error for a real-life learning problem, it is often more useful to have
an \emph{empirical error} or \emph{empirical risk} on a finite sample set $S$ of size $m$.
\begin{equation}
    L_{S}(h) = \frac{1}{m}\sum_{i=1}^{m} \loss (h(x_i), y_i)
\end{equation}

Consider a learning scenario where a learning algorithm $A$ is provided with input data of incremental size.  We use $S^i_{train}$ ($i\geq 1$) to represent the training data set in the $i$-th training round. 
In each round, the learner $A$ runs with the labeled sample $S^i_{train}$ and obtains a hypothesis 
$h_i=A(S^i_{train})$.
The test data set, $S^i_{test}$, is also sampled i.i.d. in the $i$-th round from the distribution $\D$. 
Since the distribution $\D$ is often unknown, in practice, researchers usually assess the performance of the hypothesis on the test data set as an estimator for the generalization error~\cite{shalev2014understanding}. Hence, the test error of $h_i$ over $S^i_{test}$ can be 
used to estimate $L_{\D}(h_{i})$ by $L_{S^i_{test}}(h_{i})$. 
To simplify notation, $S^i_{train}$ and $S^i_{test}$ are collectively referred to as $S^i$  throughout the remainder of the paper. 
Let $S^i$ ($i\geq 1$) be a series of training data sets drawn i.i.d. from $\D$. The plot of $L_{\D}(A(S^i))$ against the size of the training data forms a \emph{learning curve}~\cite{viering2022shape}. 

Fixing $\D$ and $\Hs$, an \emph{optimal hypothesis} has the smallest generalization error $L^*$, defined by $\arg \min_{h' \in \Hs} L_{\D}(h')$, also known as the \emph{approximation error}. 
Let $h$ be the output hypothesis of a learning algorithm, then 
and the difference between $L_\D(h)$ and $L^{*}$ 
is known as the \emph{estimation error}. Given input data $S^1, S^2, \ldots$ 
with $|S^i| < |S^{i+1}|$ for all $i\geq 1$,
one may expect $A$ to be \emph{correct} such that estimation error can be minimized 
with sufficiently large $i$, i.e., $\lim_{i\mapsto \infty}L_\D(A(S^i)) = L^{*}$.
Another important property for learner $A$, \emph{monotonicity}, has been introduced for machine learning in~\cite{loog2019minimizers,viering2019open}, which we rephrase as follows.


\begin{Definition} (locally monotonic)
Let $\D$ be the domain distribution and $S^i$ ($i\geq 1$) be a series of data sets drawn i.i.d. from $\D$. Let $A$ be a learner that generates hypothesis $h_i$ with $S^{i}$. We say $A$ is locally monotonic or $(\D, i)$-monotonic, if and only if 
\begin{equation}
    \mathop{\E}_{{S^{i+1} \sim \D}} L_{\D}(h_{i+1}) \leq \mathop{\E}_{S^{i} \sim \D} L_{\D}(h_{i}). 
\end{equation}
\label{def:LM}
\end{Definition}

\vspace{-10pt}

In this paper, we study monotonicity in the Probably Approximately Correct (PAC) learning framework. Since in general, features do not fully determine the labels of a learning target, i.e., $L^* = \arg \min_{h' \in \Hs} L_{\D}(h')\geq 0$, we study our machine learning problems within the agnostic PAC learning framework.

\begin{Definition} (Agnostic PAC Learning~\cite{shalev2014understanding}) 
    A hypothesis space $\Hs$ is agnostic PAC learnable if there exist a function $m_\Hs: (0, 1)^2 \rightarrow \Nat$ and a learning algorithm with the following property: For every $\epsilon > 0$, $\delta \in (0, 1)$ and for every probability distribution $\D$ over $\X \times \Y$, when running the learning algorithm on $m \geq m_\Hs(\epsilon, \delta)$ i.i.d. examples generated by $\D$, the algorithm returns $h \in \Hs$ such that, with probability of at least $1 - \delta$, 
    \begin{equation}
        L_{\D}(h) - L^{*} \leq \epsilon
    \end{equation}
\label{def:APL}
\end{Definition}

\vspace{-10pt}

The measure of sample complexity $m_\Hs(\epsilon, \delta)$ in agnostic PAC learning, which is polynomial in size of both $1 / \epsilon^2$ and $\log{(1 / \delta)}$, asymptotically ensures that a learner is \emph{correct} with overwhelming probability, so that $L_{\D}(A(S))$ approaches $L^{*}$ while 
the sample size $m$ (which equals $|S|$) approaches $\infty$. 
In the next section, we demonstrate that by leveraging the relationship between $m$ and $\epsilon$, the distribution 
of the estimated error of a generated hypothesis $A(S)$, where $S\sim\D^m$, can serve as 
an upper bound 
for the empirical error distribution. 
For simplicity, we will sometimes refer to the general framework as PAC learning, omitting the ``agnostic'' qualifier.

\section{PAC learning is monotone}\label{sec:pac-monotone}

In this section, we show that any effective algorithm $A$ for an agnostic PAC learnable problem is monotone. Our analysis focuses on two classes of learning problems:
(1) $\Hs$ is finite and (2) $\Hs$ has a finite VC~dimension.

\subsection{The finite hypothesis case}
\label{sec:finite-H}
According to Definition~\ref{def:APL}, if a problem is PAC learnable, the sample complexity provides a foundation for establishing a relationship between the generalization error $\epsilon$ of an output hypothesis 
$h$ from a learner $A$, the confidence level $1-\delta$, and the minimum sample size $m$. We introduce our approach as follows, 
assuming that a learning problem has %
a finite hypothesis class $\Hs$.

\begin{Lemma} (Sample Complexity for finite $\Hs$~\cite{shalev2014understanding}) 
    Let $\Hs$ be a finite hypothesis space, 
    and every training sample is taken i.i.d. from the problem distribution,
    then the class $\Hs$ is agnostic PAC learnable using the Empirical Risk Minimization (ERM) rule with the sample complexity $m_\Hs$ such that
    for any $\epsilon, \delta \in (0,1)$
    \begin{equation}
        m_\Hs(\epsilon, \delta) \leq \left \lceil \frac{2\ln(2|\Hs| / \delta)}{\epsilon^2} \right \rceil.
    \end{equation}
\label{lem:SC}
\end{Lemma}
Suppose we take a sample with size $|S| = m = \lceil \frac{2\ln(2|\Hs| / \delta)}{\epsilon^2} \rceil$, where $S\sim\D^m$, then the probability for $L_\D(A(S)) - L^{*} \leq \epsilon$ is at least $1-\delta$. That is, when $m$ is sufficiently large (we will discuss that condition in much more detail in Section~\ref{sec:experiment}), the bound representing the excess risk of the learner between the generalization loss and $L^{*}$ is less than $\epsilon$ within the confidence $1-\delta$. Given sample size $m$, we define a function $F_m: (0,1) \rightarrow (0,1)$ as the probability that an output model has generalization error bounded by $\epsilon$. 

\begin{Definition}
\label{def:Fm}
    $F_m(\epsilon) = Pr[L_{\D}(A(S)) - L^* 
    ] \leq \epsilon$, where $S\sim\D^m$ and $A$ is an ERM learner.
\end{Definition}

By taking the worst case scenario on the probability $F_m(\epsilon)$, we derive the following equation from~\cref{lem:SC}. 
\begin{equation}
\label{eq:Fm}
   F_m(\epsilon) = 1 - 2|\Hs| \cdot \exp{(-m \epsilon^2 / 2)}
\end{equation}
Since we focus on classification and use the $0$-$1$ loss, the error is necessarily bounded between $0$ and $1$. Since the minimal loss $L^*$ is determined by both the learning problem and the hypothesis class $\Hs$, we will, when context permits, assume the existence of a fixed $\mu \in [0, 1]$ such that $L^{*}=\mu$ throughout the remainder of the paper~\footnote{Formally, for $K$-class predictors, $\mu$ is bounded above by $1/K$, which represents the performance of random guessing.}. 
One may verify the edge case that when $m = \lceil\frac{2 \ln(2|\Hs| / \delta)}{\epsilon^2}\rceil$, which is its smallest possible value as per Definition~\ref{def:APL}, we have $F_m(\epsilon) = 1 - \delta$, following the PAC learnability. With a larger sample size $m$, we must have $F_m(\epsilon) > 1 - \delta$. 
Since $1 - 2|\Hs| \cdot \exp{(-m \epsilon^2 / 2)}$ is monotonic increasing for $\epsilon > 0$ and $F_m(\epsilon)$ should be non-negative, we need to add more constraints to $F_m$. The first condition is that given $F_m(\epsilon)\geq 0$, we should have $\epsilon \geq \sqrt{\frac{2\ln(2|\Hs|)}{m}}$. Therefore, we define $F_m(\epsilon) = 0$ when $\epsilon < \sqrt{\frac{2\ln(2|\Hs|)}{m}}$. 

Next, for any $m$, when $\epsilon = 1 - \mu$, we must have $F_m(\epsilon) = 1$, as the generalization error of the output of $A(S)$ trained from $S\sim\D^m$ cannot exceed $1$ by \cref{def:Fm}.
Our current choice is to define $F_m(\epsilon) = 1$ when $\epsilon \geq 1 - \mu$. The complete definition of $F_m$ is as follows. 

\begin{equation}
\begin{split}
    F_{m}(\epsilon) & = 
        \begin{cases}
        0 & \epsilon < \sqrt{\frac{2\ln(2|\Hs|)}{m}} \\
        1 - 2|\Hs| \exp{(- m \epsilon^2/2)} & \sqrt{\frac{2\ln(2|\Hs|)}{m}} \leq \epsilon < 1 - \mu \\
        1 & \epsilon \geq 1 - \mu \\
        \end{cases}.
\end{split}
\label{eq:CDF}
\end{equation}

\begin{figure*}[t]
    \centering
    \includegraphics[width = 0.32\textwidth]{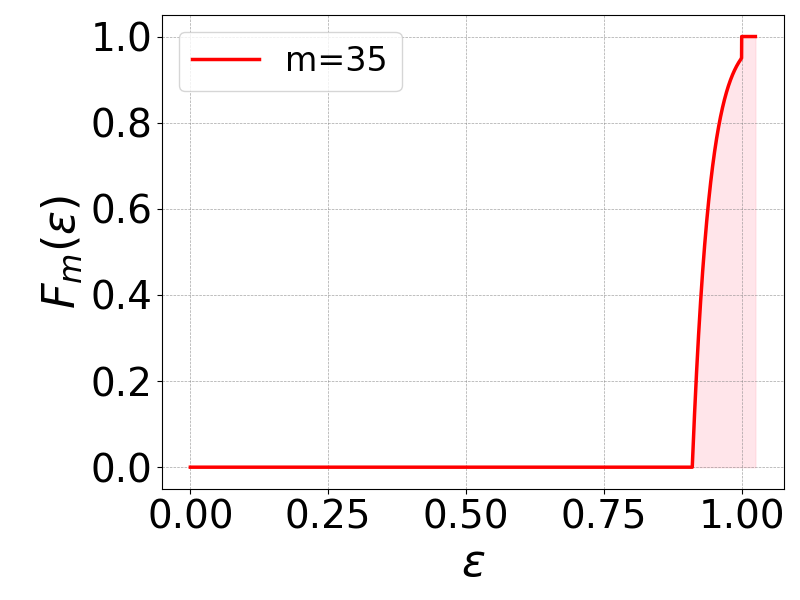}
    \includegraphics[width = 0.32\textwidth]{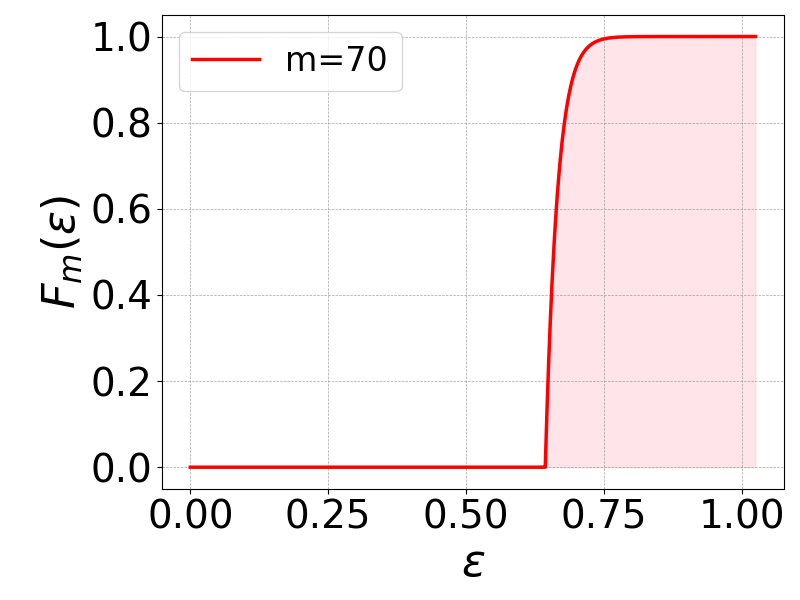}
    \includegraphics[width = 0.32\textwidth]{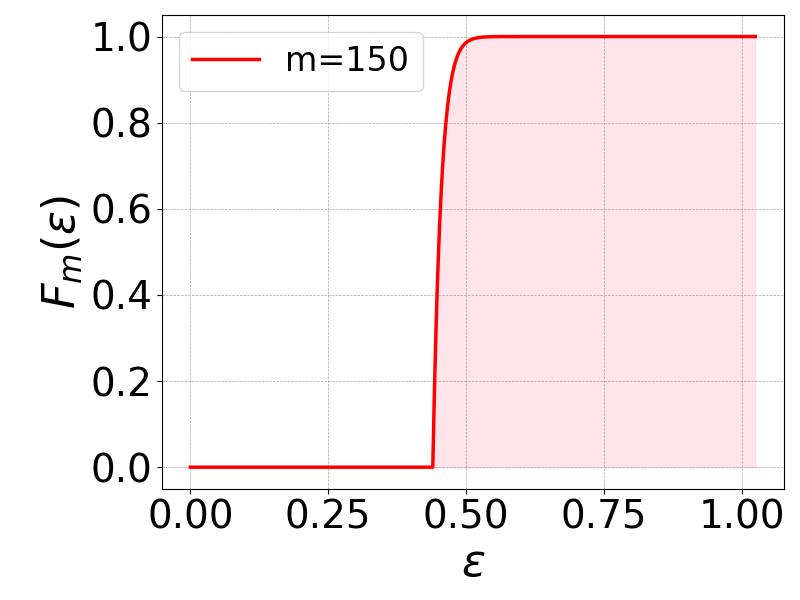}
\caption{The probability function $F_m$ with different $m$ in a finite hypothesis space.}
\label{fig:probability-1}
\end{figure*}

The three curves in \cref{fig:probability-1} assume $\mu = 0$, $|\Hs| = 1,000,000$, and $m \in \set{35, 70, 150}$.
As one may see that for the case $m = 70$, $F_m(\epsilon)$ is $1 - 1.261 \times 10^{-9}$ (already very close to $1$) when $\epsilon=1$. 

Existing approaches measure monotonicity with generalization error. That is, given a series of training data sets $S^1, S^2, \ldots$ with $|S^i| < |S^{i+1}|$ for all $i\geq 1$,
the expected error values of the generated hypotheses $A(S^i)$ are used to indicate whether the learner $A$ is locally monotonic for each $i$. We provide a more informative approach to describing monotonicity by deriving a probability density function (PDF) $f_{m}: \R \rightarrow \R$ from $F_m$.
The probability density function $f_{m}(\epsilon)$ corresponding to $F_{m}(\epsilon)$ is the solution of the following equation. 
\begin{equation}
    F_{m}(\epsilon) = \int_{-\infty}^{\epsilon} f_{m}(x) dx.
\label{eq:integral}
\end{equation}

It is worth noting that $F_{m}(\epsilon)$ exhibits partial continuity. 
Since the differential equation associated with $F_{m}(\epsilon)$ does not admit a smooth solution, the classical notion of functions on real or complex numbers will fail. 
Instead, the generalized functions provide a rigorous framework for treating discontinuous functions more like smooth functions. 

The Dirac delta function 
has since been applied routinely in physics and engineering to represent a discrete distribution, or a partially discrete, partially continuous distribution. 
Despite its name, the delta function, $\delta(x)$, is not a ``function'' in the traditional sense, but a mathematically defined distribution that acts specifically to test functions ~\cite{strichartz2003guide}.
Intuitively, $\delta(x)$ can also be loosely understood as a ``function'' on the real line which is zero everywhere except at the origin, that is, 
    $\delta(x)  = 0$ whenever $x \neq 0$,
and which is also constrained to satisfy the identity~\cite{gelfand1969generalized}
\begin{equation}
    \int_{-\infty}^{+\infty} \delta(x) dx = 1.
\end{equation}
However, a stricter definition needs to be described by its effect on the test function $f(x)$:
\begin{equation}
    \int_{-\infty}^{+\infty} \delta(x) f(x) dx = f(0).
\end{equation}

As a probability measure on $R$, $\delta(x)$ can also be characterized by its cumulative distribution function~\cite{driggers2003encyclopedia}, as follows, 
\begin{equation}
    \delta(x) = \frac{d H(x)}{d x}, 
\end{equation}
where $H(x)$ is the unit step function, defined as
\begin{equation}
    H(x) = \begin{cases} 0 & x < 0 \\ 1 & x \geq 0\end{cases}
\end{equation}

For a more detailed analysis of $F_m(\epsilon)$, it can be decomposed into two parts: $H_m(\epsilon)$, which captures the discontinuous, step-like behavior \footnote{Equivalently, $H_m(\epsilon) = 2|\Hs| \exp{(- m (1 - \mu)^2 / 2)} H(\epsilon - (1 - \mu))$.}, and $G_m(\epsilon)$, which accounts for the continuous component.
\begin{equation}
\label{eq:Hm}
    H_{m}(\epsilon) \overset{\text{def}}{=} 
    \begin{cases}
    0 & \epsilon < 1 - \mu \\
    2|\Hs| \exp{(-\frac{m (1 - \mu)^2}{2})} & \epsilon \geq 1 - \mu \\
    \end{cases}
\end{equation}
\begin{equation}
\label{eq:Gm}
    G_{m}(\epsilon) \overset{\text{def}}{=} 
    \begin{cases}
    0 & \epsilon < \sqrt{\frac{2\ln(2|\Hs|)}{m}} \\
    1 - 2|\Hs| \exp{(-\frac{m \epsilon^2}{2})} & \sqrt{\frac{2\ln(2|\Hs|)}{m}} \leq \epsilon < 1 - \mu \\
    1 - 2|\Hs| \exp{(-\frac{m (1 - \mu)^2}{2})} & \epsilon \geq 1 - \mu \\
    \end{cases}
\end{equation} 
The probability density function corresponding to the cumulative distribution function $F_{m}(\epsilon)$ is derived as follows.
\begin{equation}\label{eq:CDF-FH}
\begin{split}
    f_{m}(\epsilon) & = \frac{dF_{m}(\epsilon)}{d\epsilon} \\
    & = \frac{dG_{m}(\epsilon)}{d\epsilon} + \frac{dH_{m}(\epsilon)}{d\epsilon} \\
    & = \frac{dG_{m}(\epsilon)}{d\epsilon} + 2|\Hs| \exp{(-\frac{m (1 - \mu)^2}{2})} \frac{dH(\epsilon - (1 - \mu))}{d\epsilon} \\
    & = \frac{dG_{m}(\epsilon)}{d\epsilon} + 2|\Hs| \exp{(-\frac{m (1 - \mu)^2}{2})} \delta(\epsilon - (1 - \mu))
\end{split}
\end{equation}
where $\frac{dF_{m}(\epsilon)}{d\epsilon}$ and $\frac{dG_{m}(\epsilon)}{d\epsilon}$ respectively refer to a weak derivative of $F_{m}(\epsilon)$ and $G_{m}(\epsilon)$, and
\begin{equation}
    \frac{dG_{m}(\epsilon)}{d\epsilon} = 
    \begin{cases}
    0 & \epsilon < \sqrt{\frac{2\ln(2|\Hs|)}{m}} \; or \; \epsilon \geq 1 - \mu \\
    2|\Hs| \exp{(-\frac{m (1 - \mu)^2}{2})} & \sqrt{\frac{2\ln(2|\Hs|)}{m}} \leq \epsilon < 1 - \mu
    \end{cases}.
\end{equation}

\begin{figure}[ht]
    \centering
    \includegraphics[width = \linewidth]{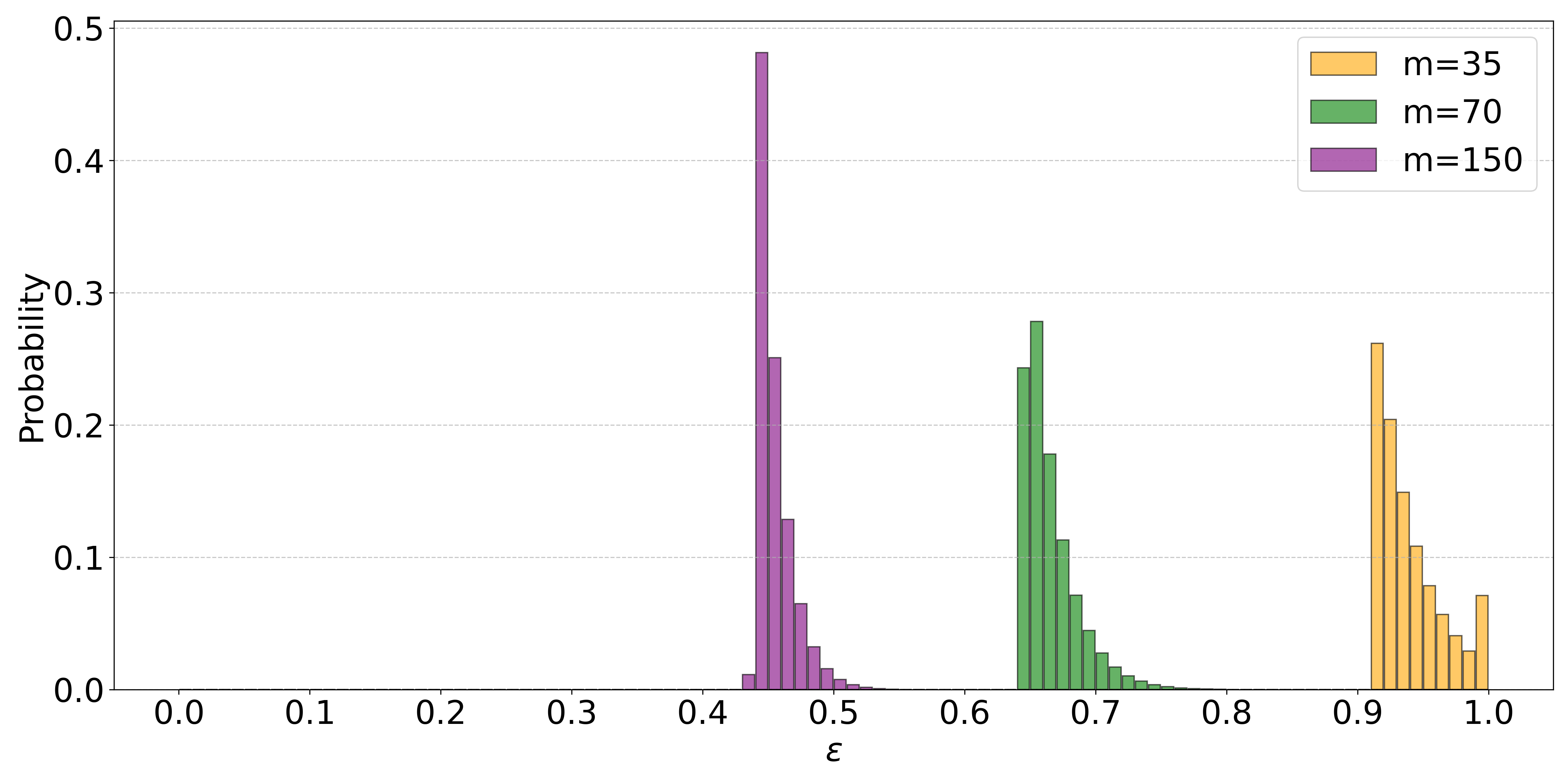}
\caption{The density function $f_m$ with different $m$ in a finite hypothesis space.}
\label{fig:probability-2}
\end{figure}

The probability density functions representing the distributions for the $F_m$ functions in \cref{fig:probability-1} are depicted in \cref{fig:probability-2}, where $|\Hs|$ is $1,000,000$, $\mu = 0$, and $m \in \{35, 70, 150\}$.

\paragraph{Realizability Assumption and tighter bound}

An agnostic PAC learnable problem is also \emph{realizable} for $\Hs$ if there exists $h'\in\Hs$ such that $L_\D(h') = 0$, or equivalently, $L^* = 0$. Moreover, this implies that $\mu = 0$ and 
we have that $L_{S}(A(S)) = 0$ for every ERM algorithm A and every training set $S$. 
This allows us to obtain a tighter bound when deriving the sample complexity for a PAC learnable problem satisfying realizability.
\begin{Lemma} (Sample Complexity for finite $\Hs$ under realizability assumption~\citep{shalev2014understanding})
    Let $\Hs$ be a finite hypothesis space, 
    and any training sample is taken i.i.d. from the problem distribution, then the class $\Hs$ is PAC learnable using the Empirical Risk Minimization (ERM) rule with sample complexity
    for any $\epsilon, \delta \in (0,1)$
    \begin{equation}
        m_\Hs(\epsilon, \delta) \leq \left \lceil \frac{\ln(|\Hs| / \delta)}{\epsilon} \right \rceil.
    \end{equation}
\label{lem:SC-FHR}
\end{Lemma}

Compared with \cref{eq:CDF}, in the following \cref{eq:CDF-FHR}, $F_{m}^{FHR}(\epsilon)$ not only fixes $\mu$ at $0$ but also 
applies a smaller sample complexity to establish the relationship among sample size $m$, accuracy parameter $\epsilon$, and confidence parameter $\delta$. 
For a finite hypothesis space under the realizability assumption, the cumulative distribution function $F_{m}^{FHR}$ is defined as follows; its corresponding density function is given in \cref{sec:FHR-D}. 
\begin{equation}
\begin{split}
    F_{m}^{FHR}(\epsilon) & = 
        \begin{cases}
        0 & \epsilon < \frac{\ln(|\Hs|)}{m} \\
        1 - |\Hs| \exp{(- m \epsilon)} & \frac{\ln(|\Hs|)}{m} \leq \epsilon < 1 \\
        1 & \epsilon \geq 1 \\
        \end{cases}.
\end{split}
\label{eq:CDF-FHR}
\end{equation}

One classic learning task over a finite hypothesis space $\Hs$ is the Boolean literal conjunction problem, which satisfies the realizability assumption. 
In \cref{sec:experiment}, we will empirically compare \cref{eq:CDF} and \cref{eq:CDF-FHR} to highlight their differences.

\subsection{The finite VC dimension case}
\label{sec:finite-VC}

The concept of sample complexity plays an important role in deriving PAC learnability of problems with infinite $\Hs$.
However, actual learning tasks are often run in infinite hypothesis spaces, for example, the decision bound in the real number system or linear separators in $R^n$ vector spaces. 
To establish learnability in such contexts, a common approach for binary classification tasks  relies on the \emph{Vapnik Chervonenkis (VC) dimension} of the hypothesis space $\Hs$, a purely combinatorial concept that quantifies the complexity of 
$\Hs$~\cite{vapnik1971uniform}.
It has been proved that an infinite hypothesis space is PAC learnable if and only if its VC dimension is finite~\cite{blumer1989learnability}, 
which denotes the size of the largest sample set that can be shattered by $\Hs$, that is, $VCdim(\Hs) = \max\{m: \Pi_{\Hs}(m) = 2^m\}$.
In fact, $\Pi_{\Hs}(m)$ is the maximum number of distinct ways in which $m$ points can be classified using hypotheses in $\Hs$.
An important conclusion can be drawn by using $\Pi_{\Hs}(m)$ to estimate the relationship between generalization error and $L^{*}$:
\begin{Corollary}
\label{cor:CI-VC}
    For a hypothetical space $\Hs$, $h \in \Hs$, $m \in \Nat$, and $0 < \epsilon < 1$. When $m > 2/\epsilon^2$, we have that
    \begin{equation}
        \P(L_{\D}(h) - L^{*} > \epsilon) \leq 4 (\frac{2em}{d})^d \exp{(- \frac{m \epsilon^2}{32})},
    \end{equation}
\end{Corollary}
A more detailed explanation of \cref{cor:CI-VC} can be found in \cref{sec:SC-VCAD}. 
The sample complexity for VC dimension we are concerned about is defined as follows:
\begin{Corollary} (Sample Complexity for VC dimension)
    Assume the VC dimension of the hypothesis space $\Hs$ is $d$, and any training sample is taken i.i.d. from the problem distribution. 
    Then, the class $\Hs$ is PAC learnable using the ERM rule with sample complexity
    \begin{equation}
        m_\Hs(\epsilon, \delta) \leq \left \lceil \frac{32d\ln(2em/d) + 32\ln(4 / \delta)}{\epsilon^2} \right \rceil.
    \end{equation}
\label{lem:SC-VC}
\end{Corollary}
Similarly to the finite $\Hs$ case, we derive a probability distribution, $F_{m}^{VC}$, as a performance lower bound for any PAC learnable $\Hs$ with VC dimension $d$ and sample size $m$. We are able to derive the probability density function in a similar way to the finite $\Hs$ case. For the sake of readability, we leave our derivation procedure in \cref{sec:VCAD}.

\begin{equation}
\begin{split}
    F_{m}^{VC}(\epsilon) & = 
        \begin{cases}
        0 & \epsilon < \sqrt{\frac{32 d \ln{(2em/d)} + 32\ln{(4)}}{m}} \\
        1 - 4(\frac{2em}{d})^d \exp{(\frac{- m \epsilon^2}{32})} & \sqrt{\frac{32 d \ln{(2em/d)} + 32\ln{(4)}}{m}} \leq \epsilon < 1 - \mu \\
        1 & \epsilon \geq 1 - \mu \\
        \end{cases}.
\end{split}
\label{eq:CDF-VC}
\end{equation}

\section{Experiment}\label{sec:experiment}

As we have conjectured, given a PAC learnable problem with any $m>0$, the probability density function $f_m$ serves as an upper bound for the generalization loss distribution of the hypotheses generated by an ERM learning algorithm with training sets of size $m$. To support our conjecture, we conducted systematic evaluations on three distinct PAC learnable problems. For each candidate algorithm, performance was assessed by collecting a \emph{distribution} of loss values across output models, rather than relying solely on a single average metric, using fixed training sample sizes. Although the learning algorithms are deterministic, the sampling process is probabilistic in all examined learning problems. Finally, we compare the empirical performance distributions of the generated models with the theoretical distributions predicted by PAC theory.
The results have confirmed that the theoretical distributions effectively merge into the experimental distributions when the sample size is sufficiently large. Moreover, both learning curves generated from the series of distributions appear monotone, where the monotonicity of the theoretical distributions is guaranteed by definition.

\subsection{Learning Problems in the Experiment}

We use the Boolean literal conjunction problem to 
showcase the result for the class of learning problems with finite hypotheses $\Hs$ (\cref{example:1}). The case of finite VC dimensions was explained using the threshold function learning problem (\cref{example:2}). It is noted that both 
\cref{example:1} and \cref{example:2} are under the \emph{realizability} assumption. 
For the class of (non-realizable) agnostic learning with finite VC dimension,
we analyze the Iris classification problem (\cref{example:3}). 

\begin{Example} (Conjunction of boolean literals)
Consider a finite set of propositions $P=\set{x_1, x_2,\ldots x_n}$. A boolean literal is either a proposition $x_i$ or its negation $\bar{x}_i$, and the task is to learn a formula that is a conjunction of boolean literals from a set of positive and negative examples. Given a formula $\phi\in\Hs$ and for each $x_i\in P$, $x_i$ may appear positive or negative in $\phi$, or $x_i$ does not appear in $\phi$. A training example is a mapping from $P$ to $\set{0, 1}$, and the example is positive if it is consistent with the target formula. Otherwise the example is negative. Suppose $n=5$ and let a target formula be $\bar{x}_1 \wedge x_3 \wedge x_5$, then $(0,1,1,0,1)$ is a positive training example. This problem has a finite hypothesis space $|\Hs| = 3^n$ given fixed $n$.
\label{example:1}
\end{Example}

Learning a conjunction of boolean literals has been shown PAC learnable~\cite{valiant1984theory}. Intuitively, a positive example $(0, 1, 1, 0, 1)$ implies that the target concept cannot contain the literals $\bar{x}_2$, $\bar{x}_3$ and $\bar{x}_5$ and that it cannot contain the literals $x_1$ and $x_4$. 
In contrast, a negative example is not as informative, as it does not indicate which of its $n$ bits are incorrect.\footnote{As an example, given a negative case $(1,1,1,1,1)$, it does not provide information on which of the five literals should be $0$.}
A simple algorithm for finding a consistent hypothesis is presented as follows, based only on positive examples~\cite{mitchell1982generalization}.
\begin{enumerate}
    \item Let the hypothesis formula $\phi_h$ initially be $x_1\wedge\bar{x}_1\wedge x_2\wedge \bar{x}_2 \dots x_n\wedge\bar{x}_n$
    \item For each positive example $(b_1, ..., b_n)$ and $i = 1\dots n$, if $b_i = 1$ then $\bar{x}_i$ is ruled out from $\phi_h$; If $b_i = 0$ then $x_i$ is ruled out from $\phi_h$.
    \item The formula consisting of all the literals not ruled out in the end is a hypothesis 
    generated by the algorithm with the training sample.
\end{enumerate}
We compare the theoretical performance lower bound with the result distribution on solving instances of learning conjunction of boolean literals using this algorithm later in this section. 

\begin{Example}(Threshold function)
A threshold function is a real value that separates the real line (of infinite values) into two subsets.
Let $\Hs$ be the set of threshold functions, then $\Hs = \set{h_a: a \in R}$, where $h_a: R \rightarrow \{0, 1\}$ is a function such that $h_a(x) = \mathbb{I}_{[x < a]}$. 
In other words, $\mathbb{I}_{[x < a]}$ is $1$ if $x < a$ and $0$ otherwise. 
A training sample can be a set of pairs $(x, y)$ where $x\in\R$ and $y\in\set{1,0}$ indicating $h_a(x) = y$ for a learning target $a$.
\label{example:2} 
\end{Example}
It has been proved that the VC dimension of the threshold function is $1$, so the problem is PAC learnable~\cite{blumer1989learnability}. A candidate algorithm is to take the value $a$ as the \emph{smallest} $x$ where $(x, y)$ is in the training set with $y=0$. It is easy to see that a larger training set size increases the probability of obtaining a value that is sufficiently close to the learning target. 

\begin{Example} (Iris classification)
This is a classic data set from Fisher~\cite{fisher1936use}, and widely used in statistics and machine learning. 
The data set contains three iris flower species, each represented by four features: sepal length, sepal width, petal length, and petal width. Notably, Iris Versicolor and Iris Virginica are not linearly separable. 
\label{example:3} 
\end{Example}

In this task, we focus on classifying Iris Versicolor and Iris Virginica only, as VC dimension primarily applies to binary classification problems~\footnote{In multi-class classification settings, the VC dimension is commonly extended by the Natarajan dimension~\cite{natarajan1989learning}.}. 
Since the VC dimension of the class of nonhomogenous halfspaces in $\R^d$ is $d + 1$, we may regard the VC dimension of the support vector machine (SVM) algorithm on the Iris data set as $5$.

\subsection{Experiment}

\paragraph{Experiment setup.} 
We run the experiment on Windows 11 equipped with Intel Core i7-14700 2.10GHz processor, 32G RAM, and Nvidia GeForce GTX4070. 
%
\begin{algorithm}[t]
\footnotesize
\caption{Getting an empirical error distribution from $k$ samples of size $m$}
\begin{algorithmic}[1] 
    \REQUIRE the size of training sets $m$, the number of iterations $k$, and the number of intervals $l$ \\
    \ENSURE the probability density function $P_m$ \\
    \STATE A hypothesis is randomly generated, defined as \emph{GT} (ground truth).
    \STATE $R = \{\}$.
    \FOR {$i = 1, 2, ..., k$}
        \STATE $m$ instances are i.i.d. sampled and their labels are determined by using \emph{GT}. 
        \STATE A hypothesis $h$ is generated by running a learning algorithm on the $m$ instances. 
        \STATE Calculate the generalization error 
        between $h$ and \emph{GT}, $R_{i}$.
        \STATE Append to $R$: $R \leftarrow [R; R_{i}]$
    \ENDFOR
    \STATE The probability density function $P_m$ is obtained by calculating the frequency of 
    values in $R$ across $l$ intervals.
    \STATE \textbf{return} $P_m$
\end{algorithmic}
\label{alg:EDF}
\end{algorithm}

For each learning problem, given sample size $m$, we generate a collection of $k$ models, each trained from a distinct sample, and apply a validation procedure to estimate the generalization error of each generated model. 
Instead of generating an average performance from collected data, we keep all 
errors that form an empirical error distribution
, denoted as $P_{m}$. 
The entire process can be described in Algorithm~\ref{alg:EDF}. 
Note that line 5 of \cref{alg:EDF} describes only an abstract training process; for specific learning algorithms, see \cref{example:1}, \cref{example:2}, and \cref{example:3}. 
In line 6 of \cref{alg:EDF}, the generalization error is directly computed by measuring the proportion of misclassified examples over the 
the sample space against GT (ground truth).
Throughout this study, we set $k = 1000$, $l = 100$, and we assume that the data for both learning problems are sampled from uniform distributions. 

A theoretical distribution can be generated independently~\footnote{These distributions are generated from \cref{eq:CDF-FHR} (the case of finite $\Hs$) and \cref{eq:CDF-VC} (the case of finite VC).}.
Taking \cref{example:1}, suppose a Boolean literal conjunction learning problem has $10$ boolean variables, and the sample size $m=25$.
Since each literal can be included with or without negation, or not included, we have $|\Hs|=3^{10}=59,049$. By instantiating \cref{eq:CDF-FHR}, we can get 
\[
\begin{split}
    F_{m}^{FHR}(\epsilon) & = 
        \begin{cases}
        0 & \epsilon < 0.439 \\
        1 - 59,049 \times \exp{(- 25 \cdot \epsilon)} & 0.439 \leq \epsilon < 1 \\
        1 & \epsilon \geq 1 \\
        \end{cases}.
\end{split}
\]
Similar to $P_{m}$, we discretize the entire error range, thereby obtaining the theoretical error distribution $Q_{m}$. 

\begin{figure*}[!ht]
    \centering
    \subfloat{
        \includegraphics[width = 0.32\textwidth]{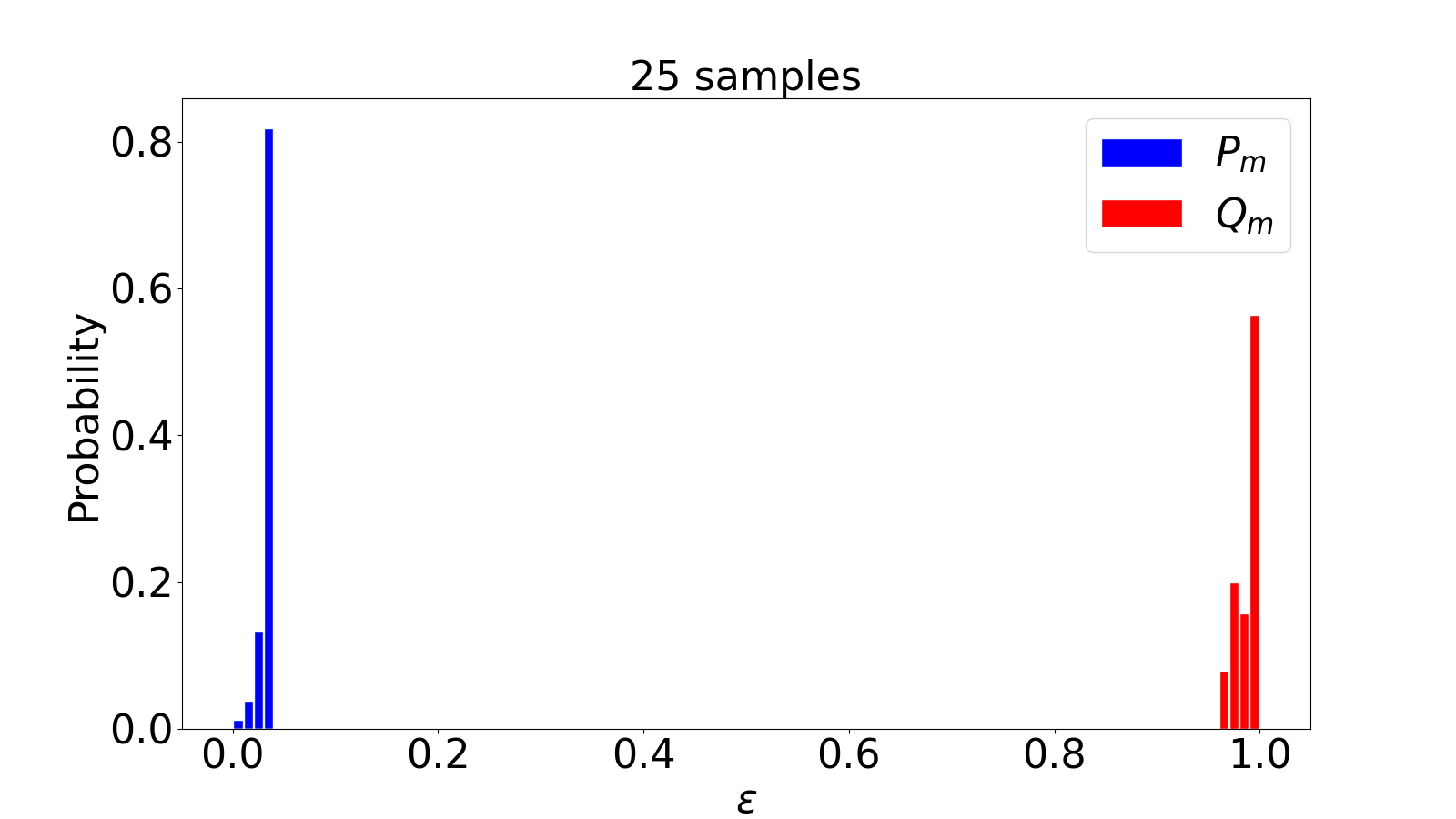}
    }
    \hfill
    \subfloat{
        \includegraphics[width = 0.32\textwidth]{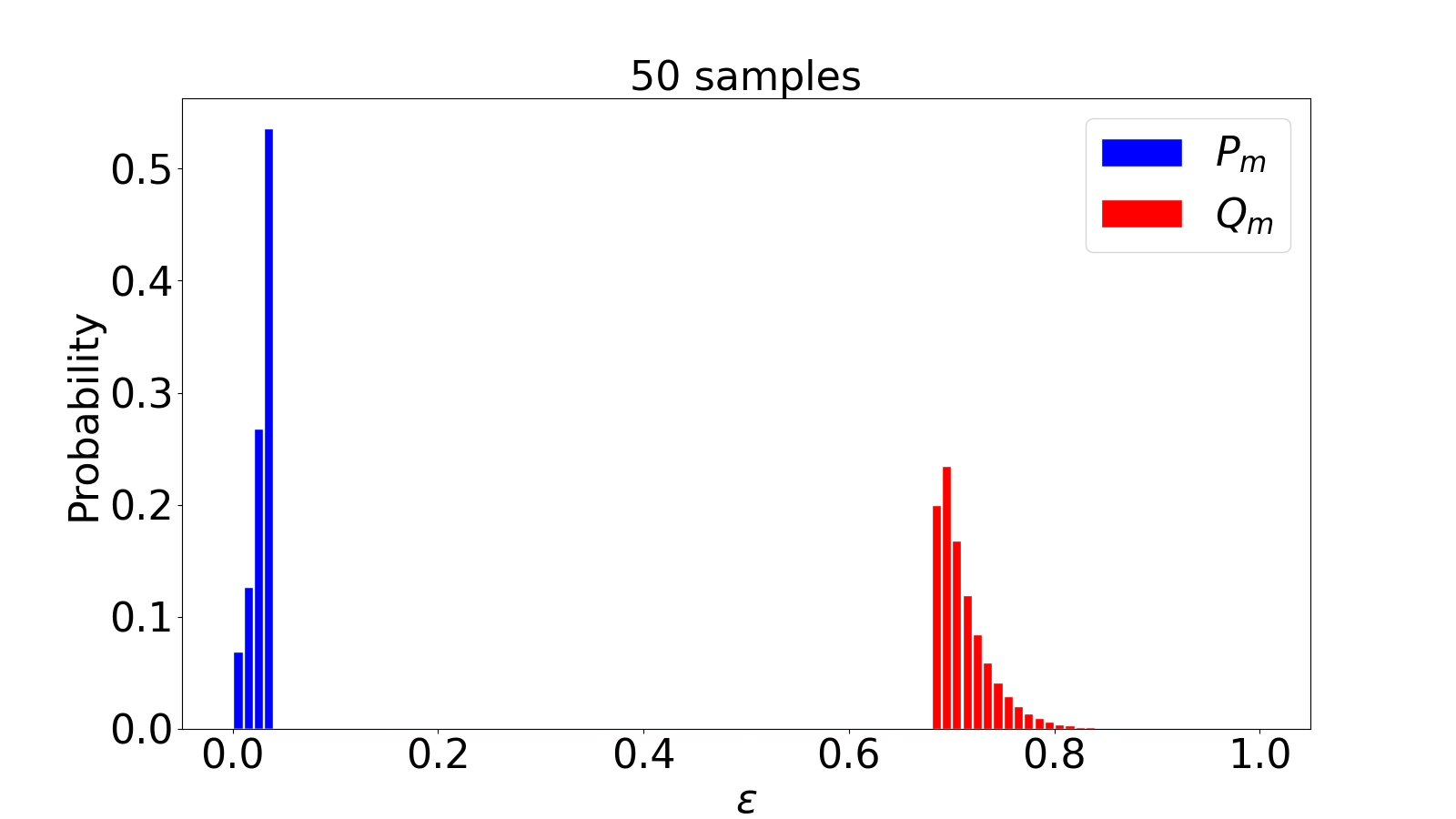}
    }
    \hfill
    \subfloat{
        \includegraphics[width = 0.32\textwidth]{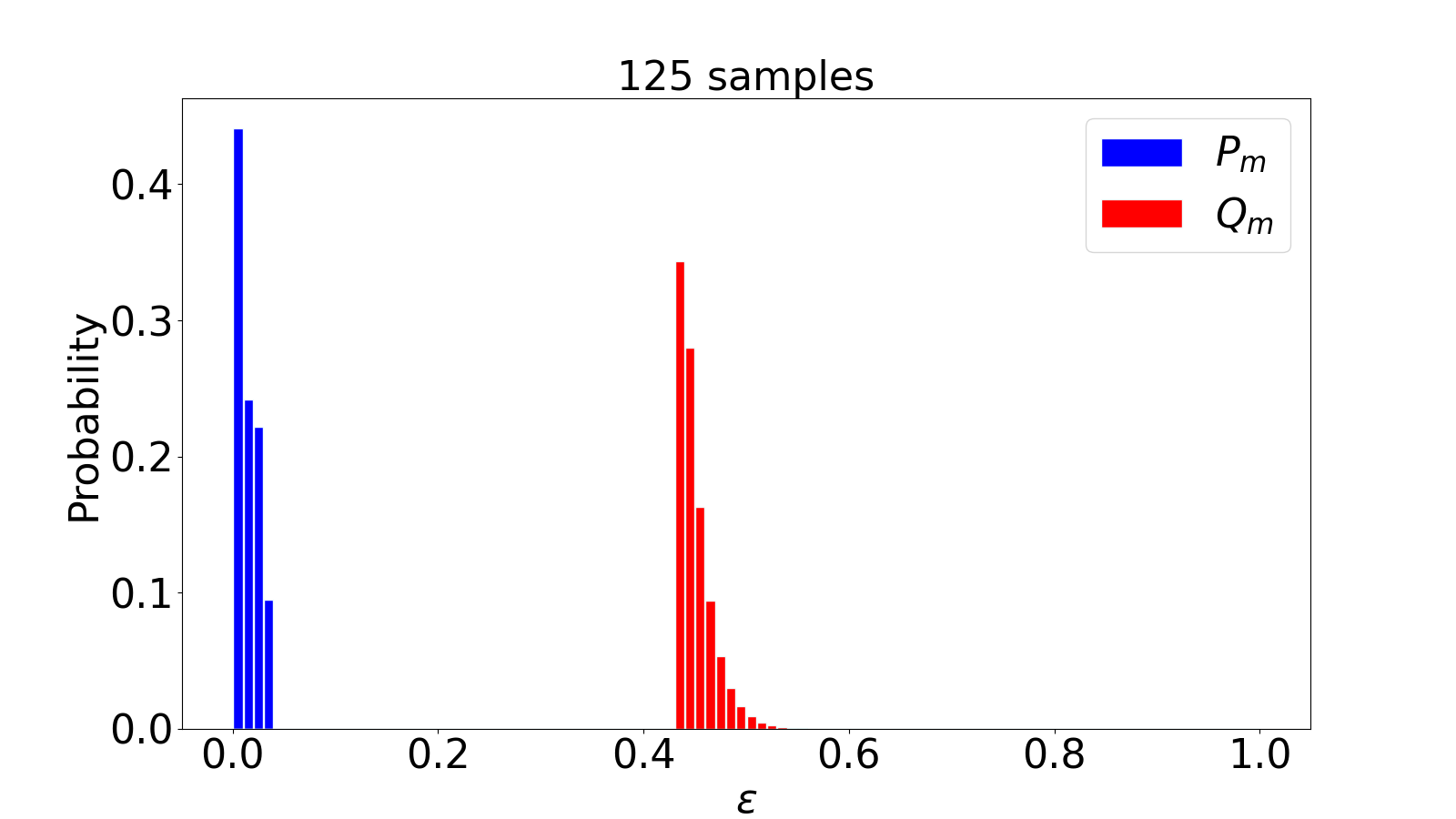}
    }
    \\
    \subfloat{
        \includegraphics[width = 0.32\textwidth]{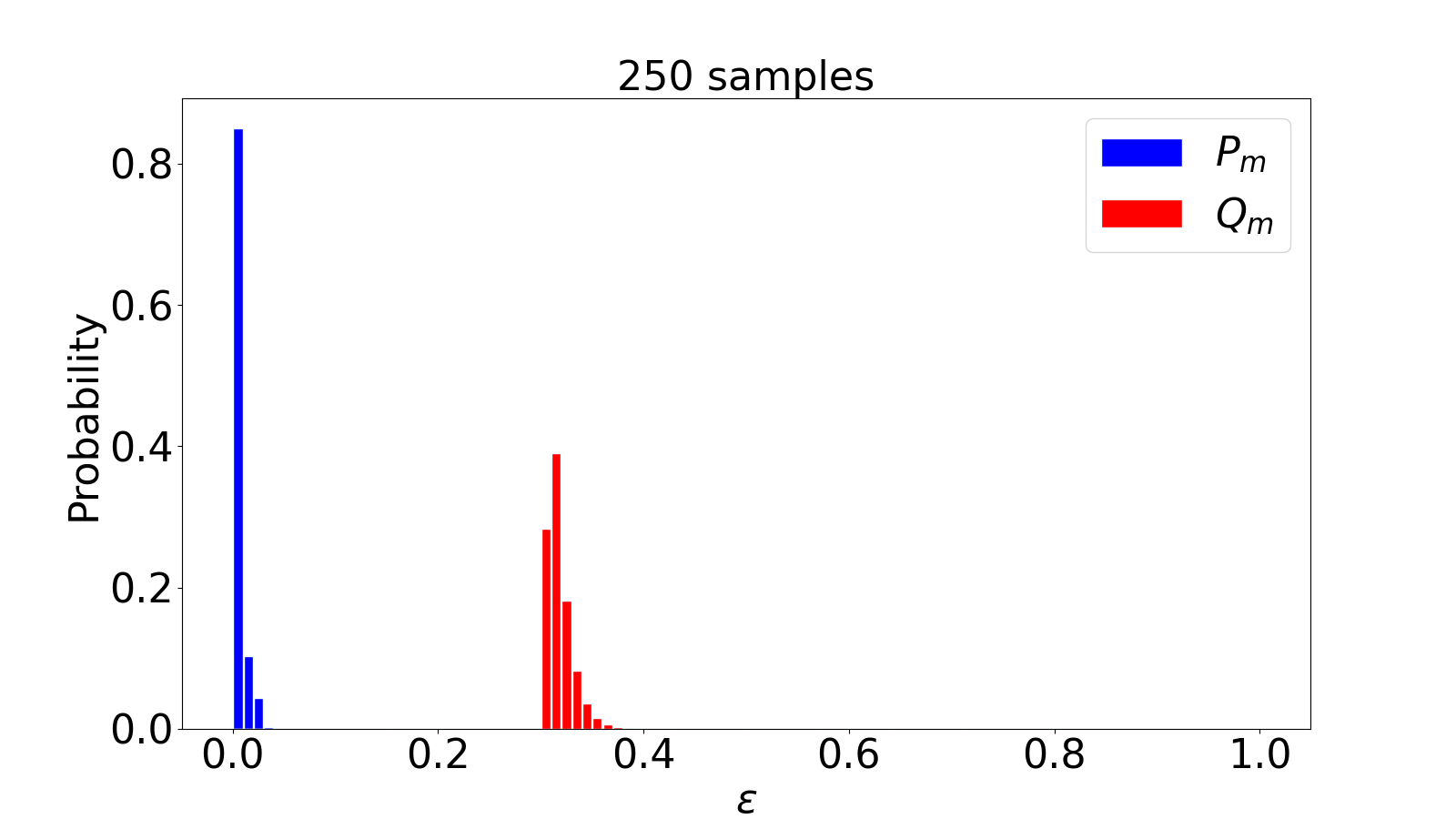}
    }
    \hfill
    \centering
    \subfloat{
        \includegraphics[width = 0.32\textwidth]{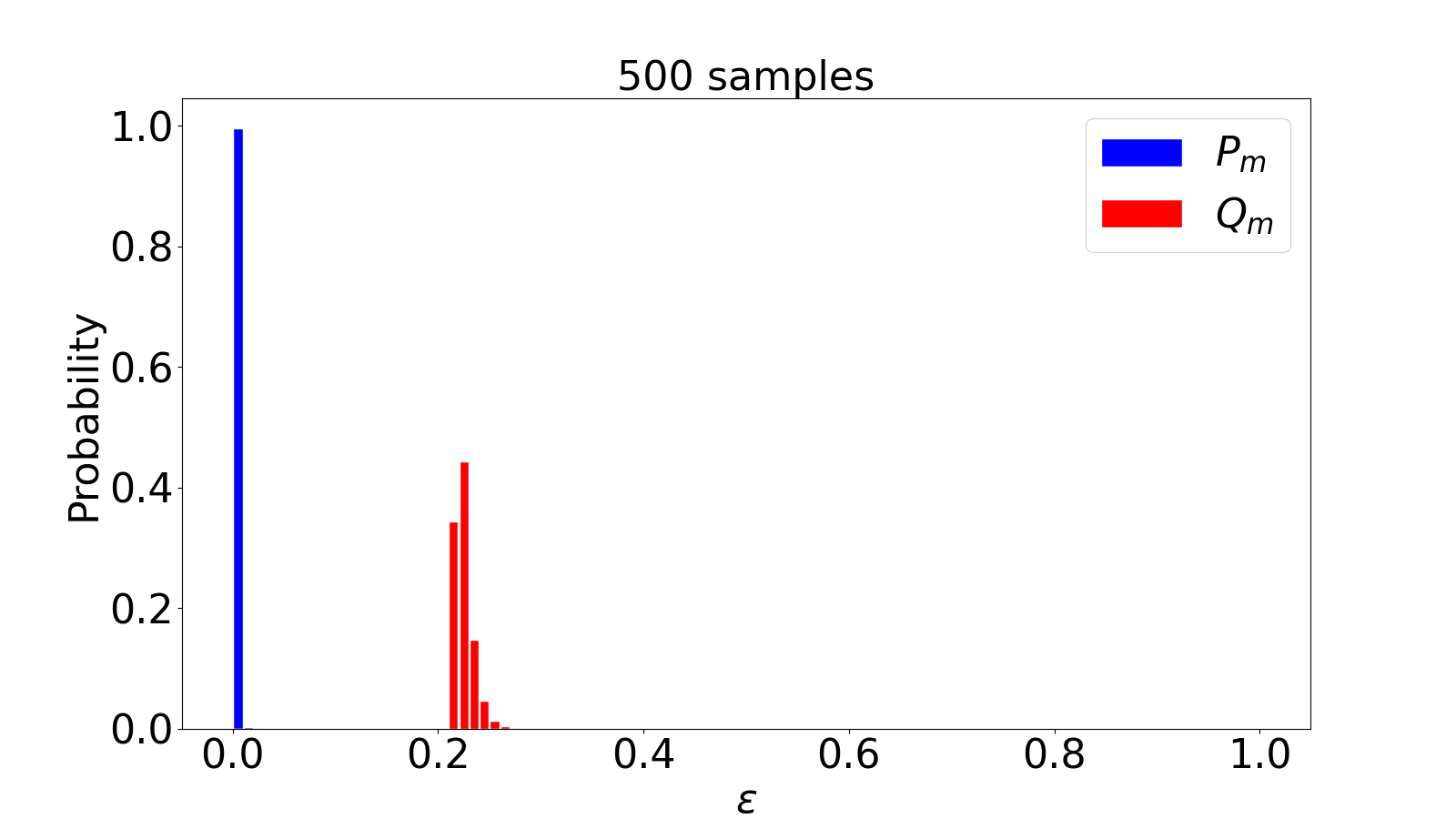}
    }
    \hfill
    \subfloat{
        \includegraphics[width = 0.32\textwidth]{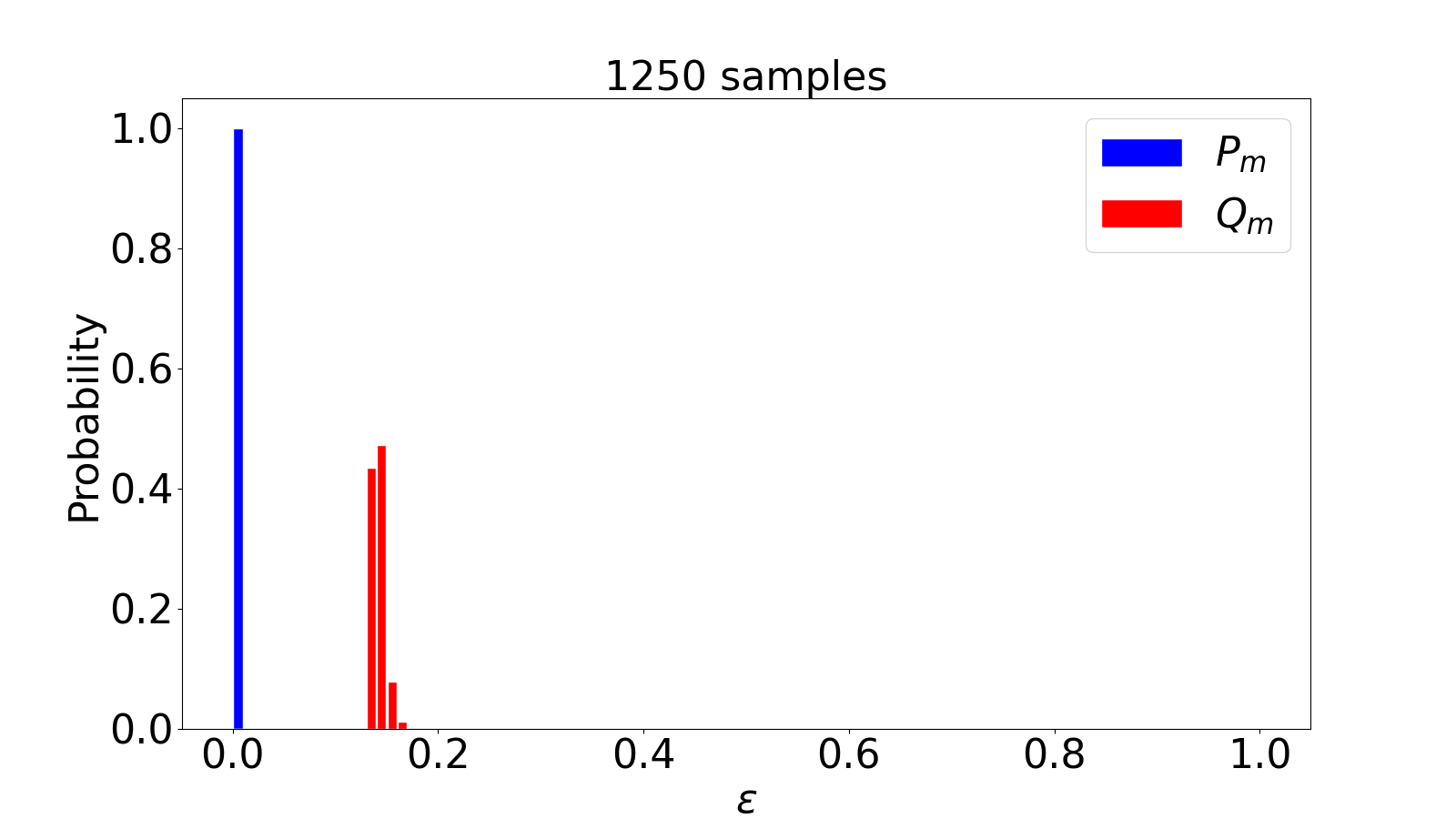}
    }
\caption{Distributions on the Boolean literal conjunction learning problem with different sample size $m$.}
\label{fig:CBL}
\end{figure*}

\begin{figure*}[!ht]
    \centering
    \subfloat[The mean.]{
        \includegraphics[width = 0.32\linewidth]{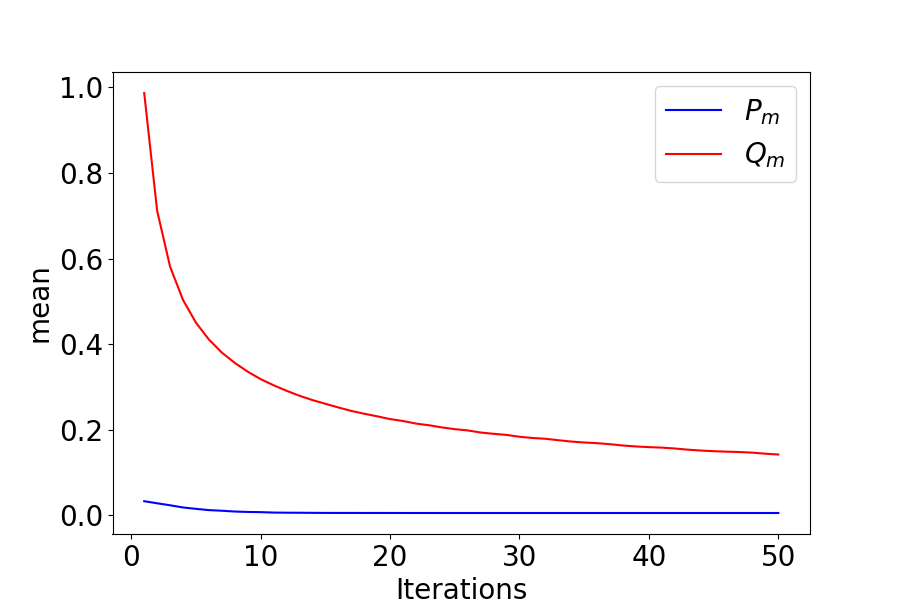}
    }
    \hfill
    \subfloat[The standard deviation.]{
        \includegraphics[width = 0.32\linewidth]{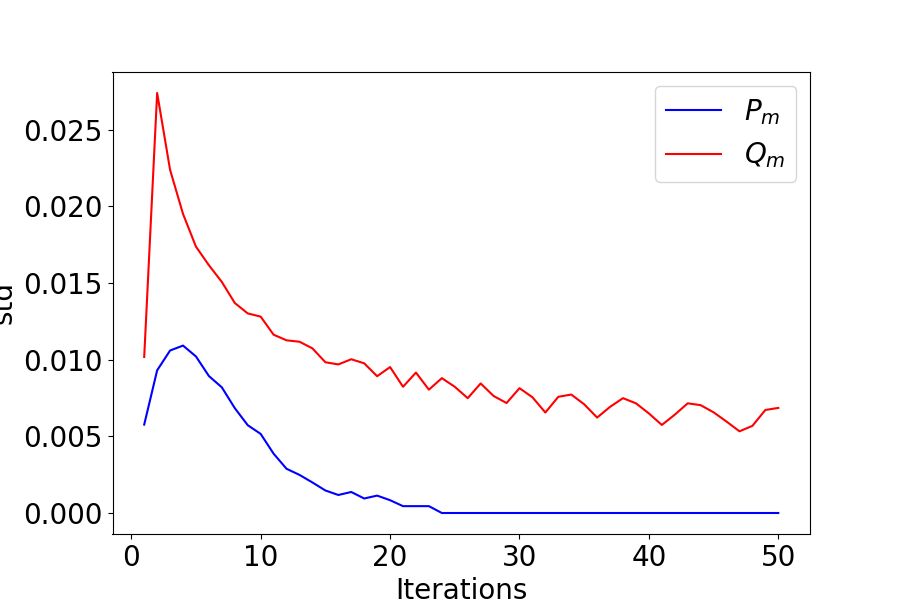}
    }
    \hfill
    \subfloat[The Wasserstein distance.]{
        \includegraphics[width = 0.32\linewidth]{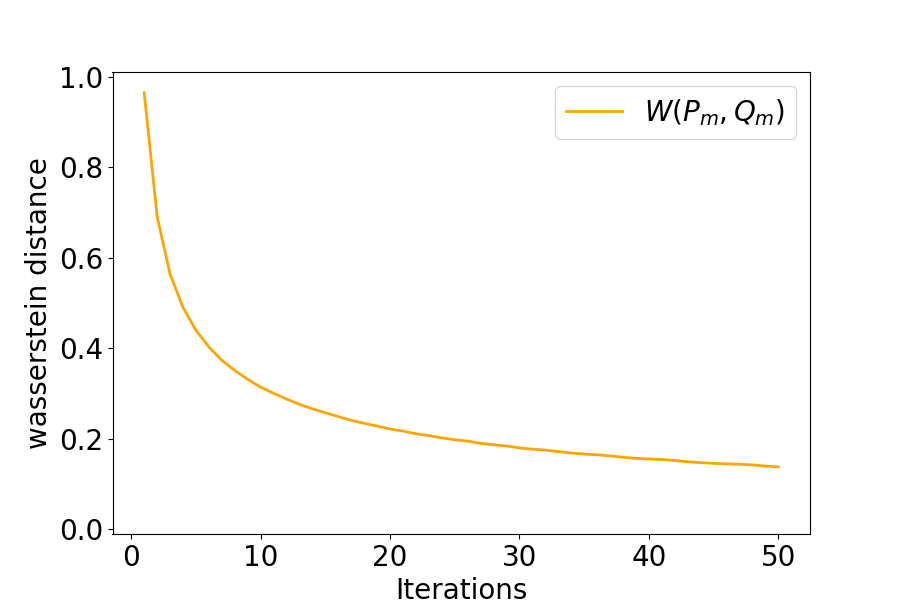}
    }
\caption{Quantitative analysis of the Boolean literal conjunction learning problem: (a) mean, (b) standard deviation, and (c) Wasserstein distance. 
}
\label{fig:CBL_QA}
\end{figure*}

As we design the experiment, we pose the following two research questions which will be addressed in our analysis of the experiment data.

\begin{itemize}
    \item[RQ1] For the two PAC learnable problems, with an increase of $m$, do the measured error distributions for the trained models exhibit a monotonic decrease?
    \item[RQ2] Can we regard the theoretical distribution as a conservative estimate for the empirical distribution, especially regarding monotonicity in learning?
\end{itemize}

\paragraph{Experiment results.}


\begin{figure*}[!ht]
    \centering
    \subfloat{
        \includegraphics[width = 0.32\textwidth]{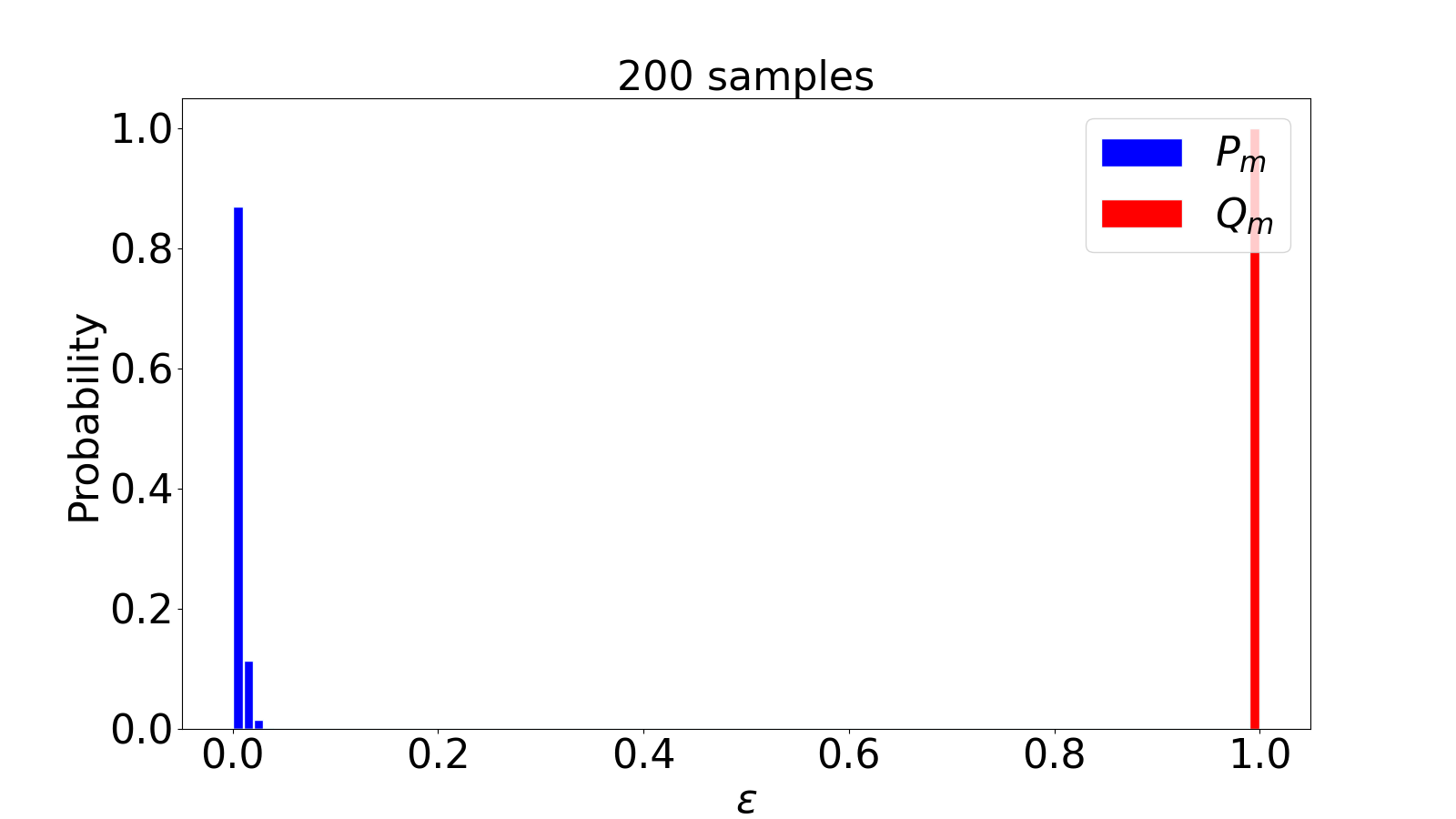}
    }
    \hfill
    \subfloat{
        \includegraphics[width = 0.32\textwidth]{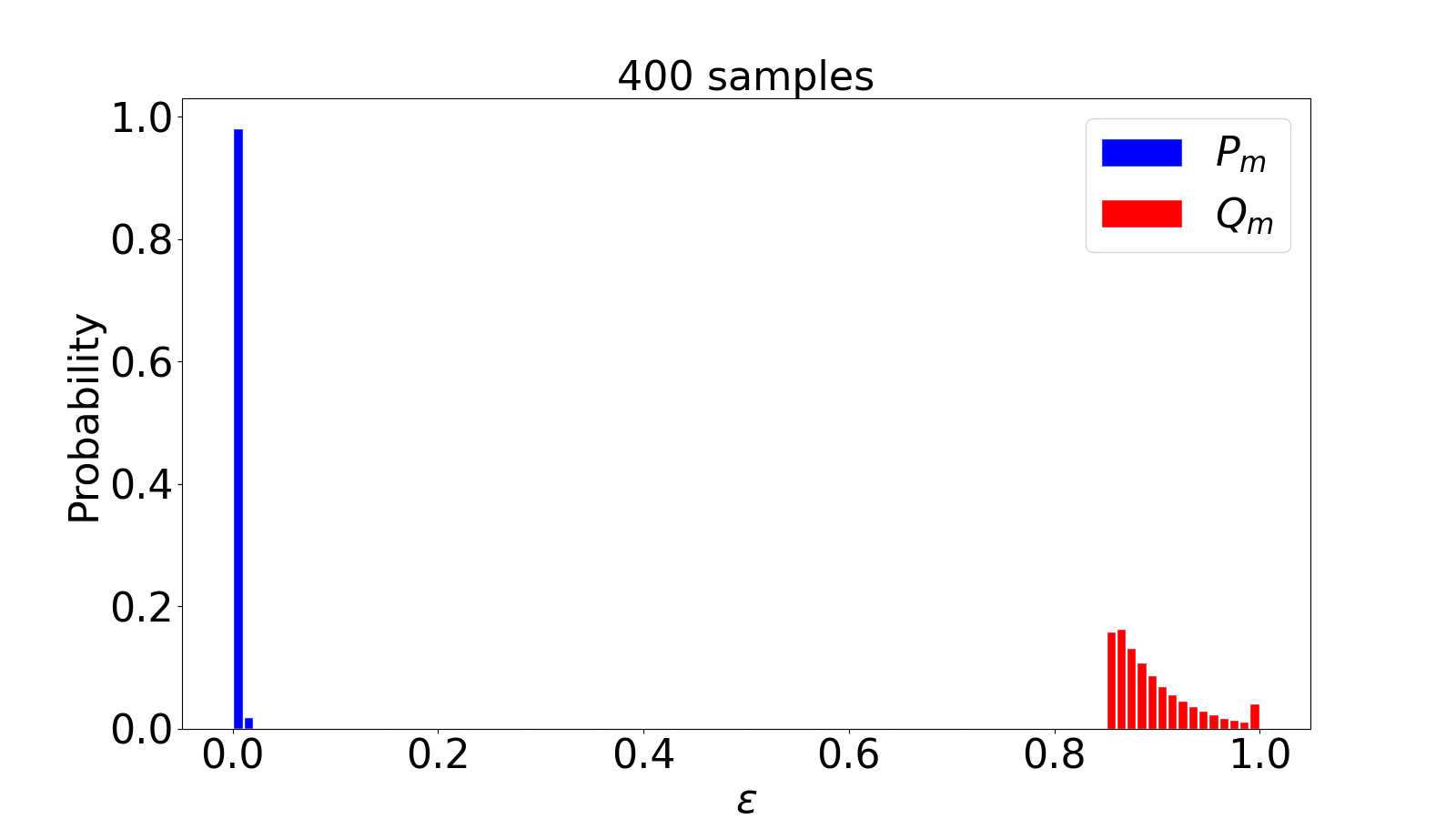}
    }
    \hfill
    \subfloat{
        \includegraphics[width = 0.32\textwidth]{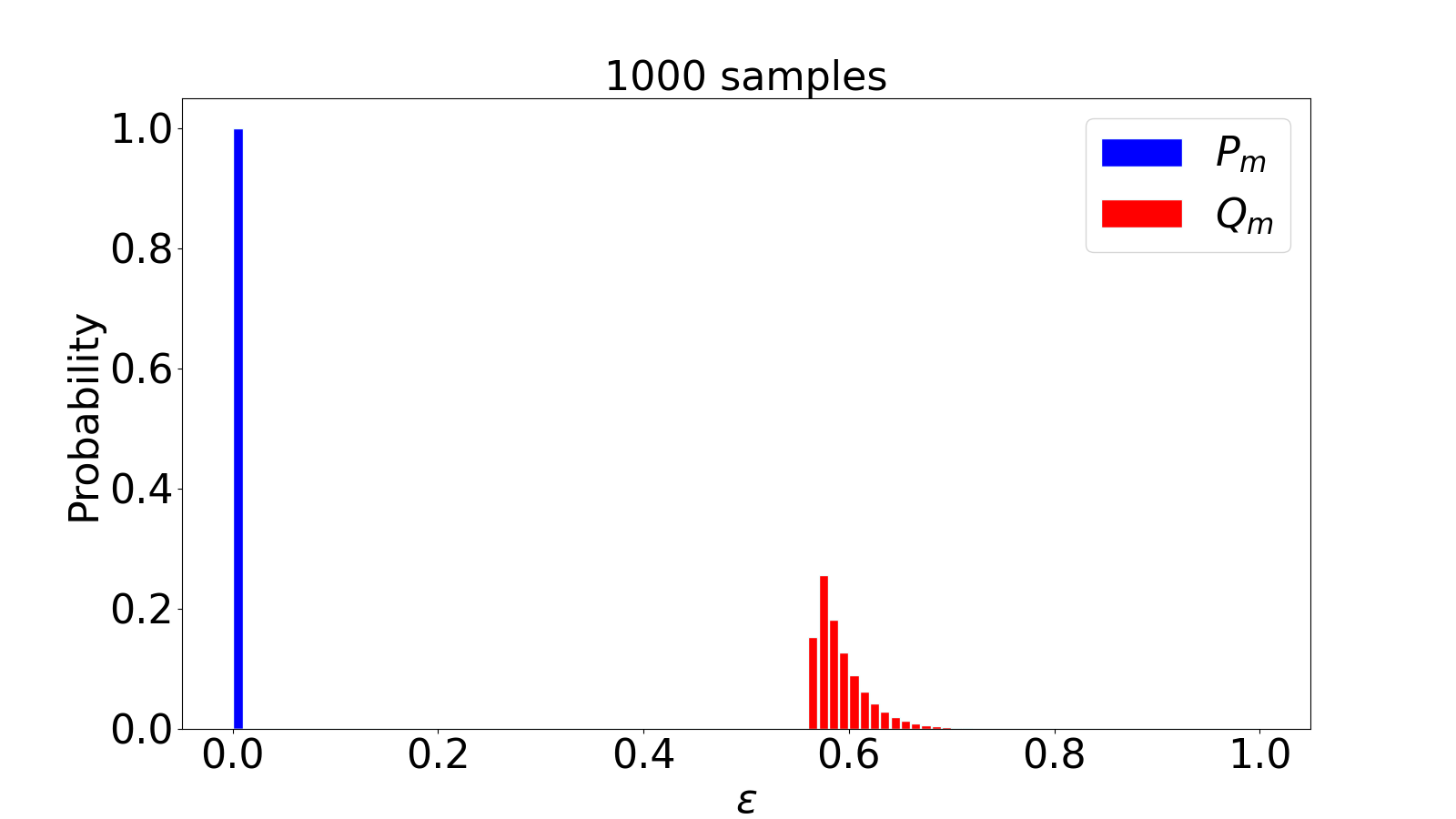}
    }
    \\
    \subfloat{
        \includegraphics[width = 0.32\textwidth]{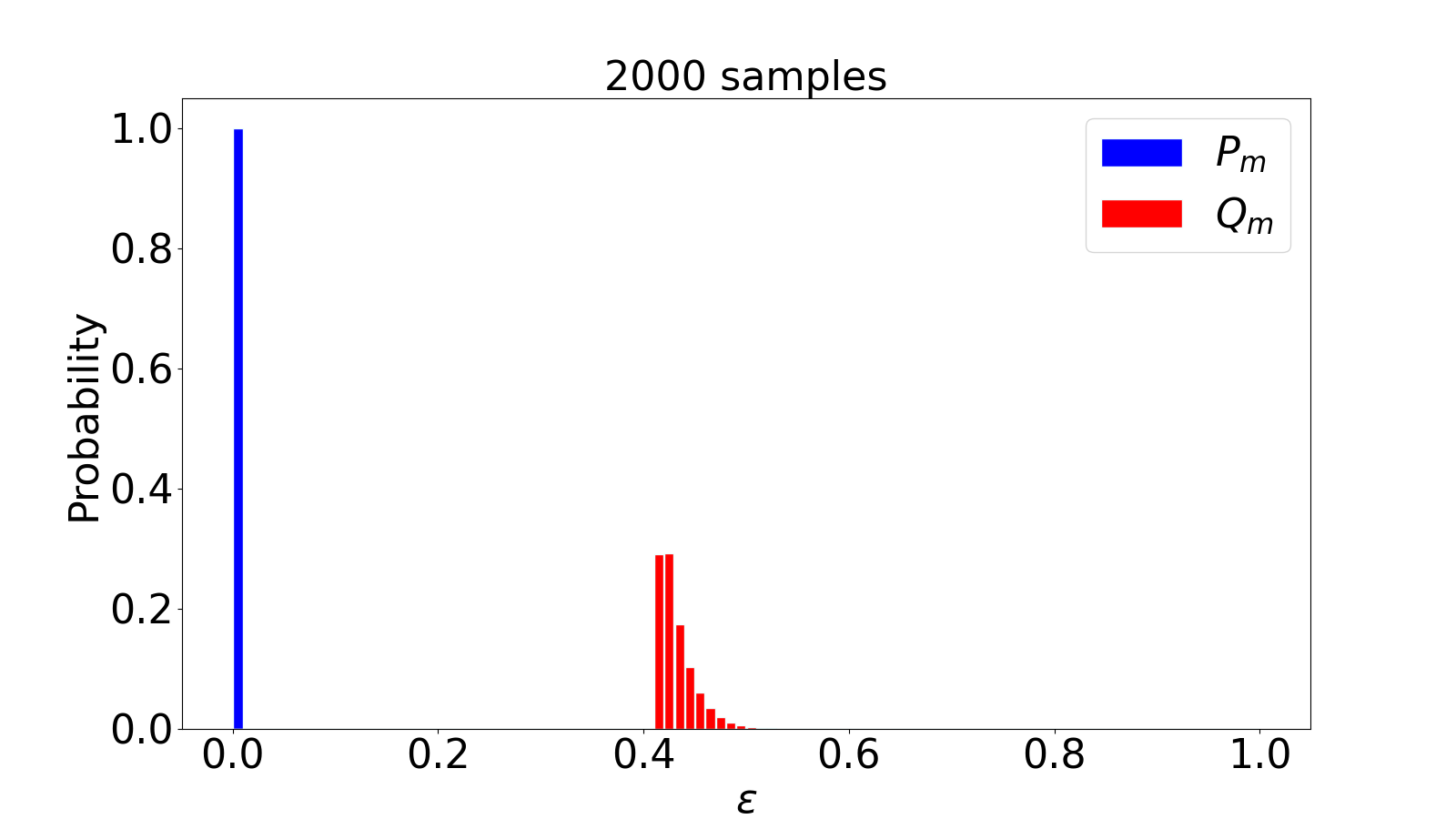}
    }
    \hfill
    \subfloat{
        \includegraphics[width = 0.32\textwidth]{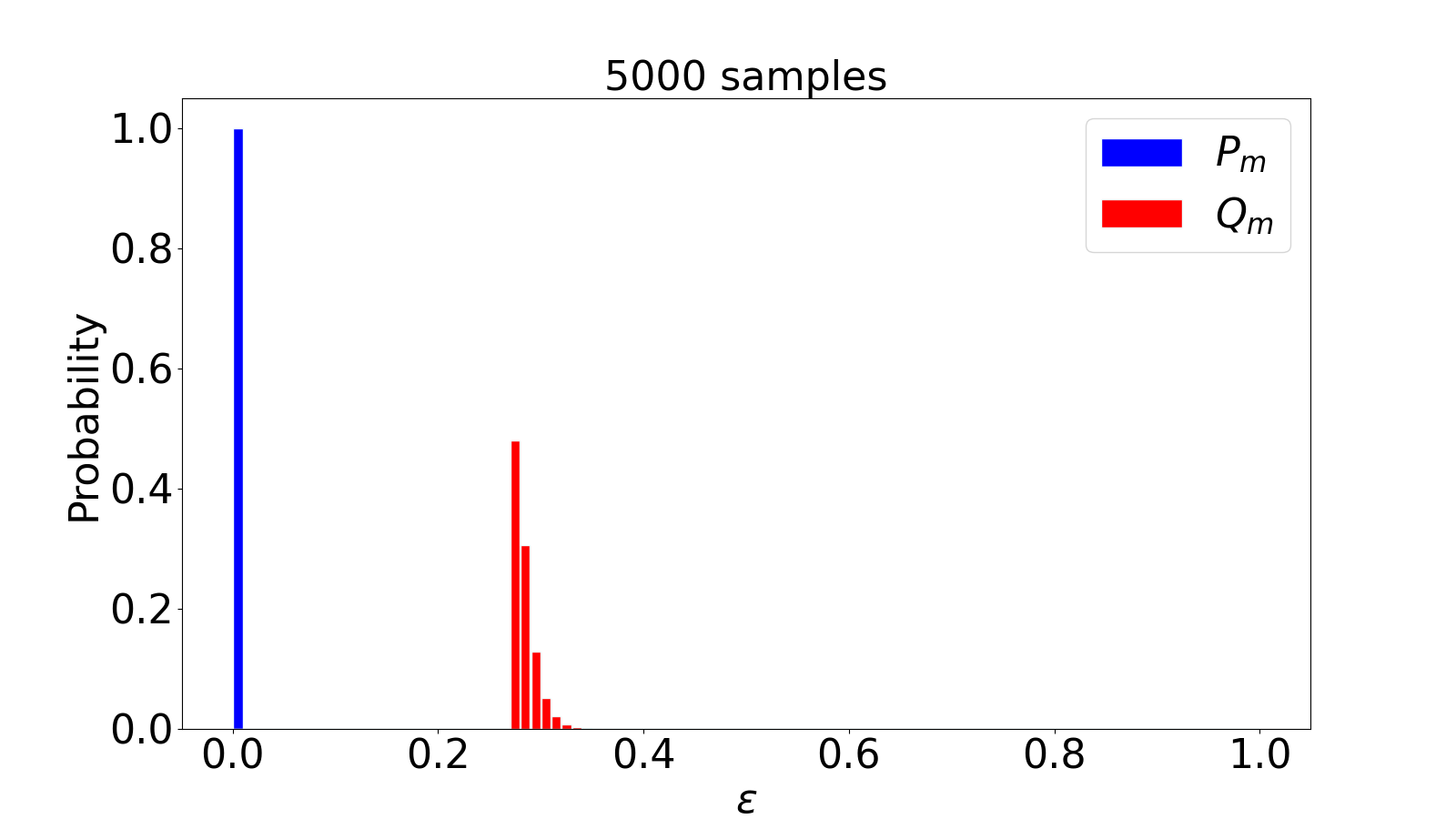}
    }
    \hfill
    \subfloat{
        \includegraphics[width = 0.32\textwidth]{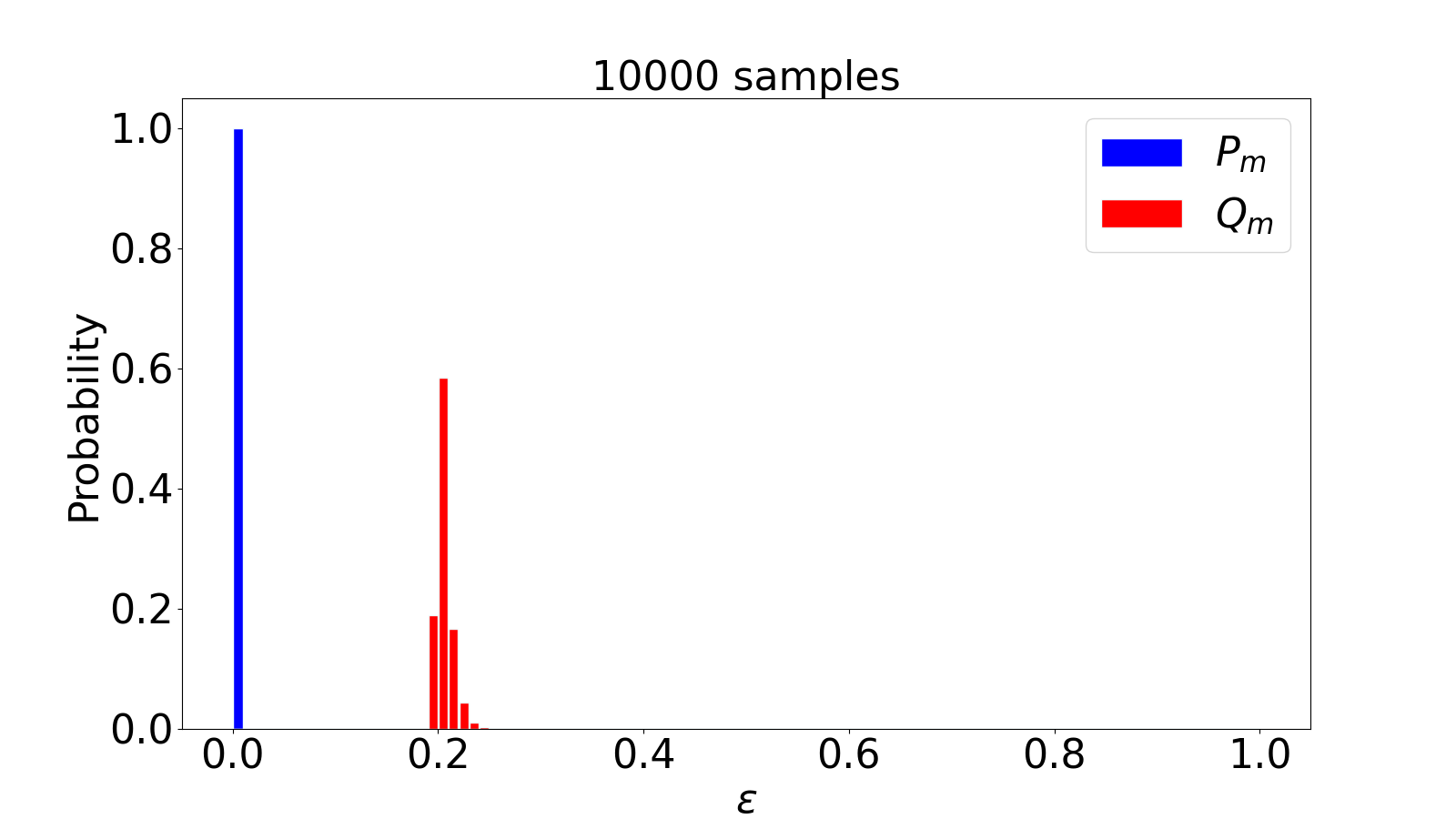}
    }
\caption{Distributions on the threshold function learning with different sample size $m$.}
\label{fig:TH}
\end{figure*}

\begin{figure*}[ht]
    \centering
    \subfloat[The mean.]{
        \includegraphics[width = 0.32\linewidth]{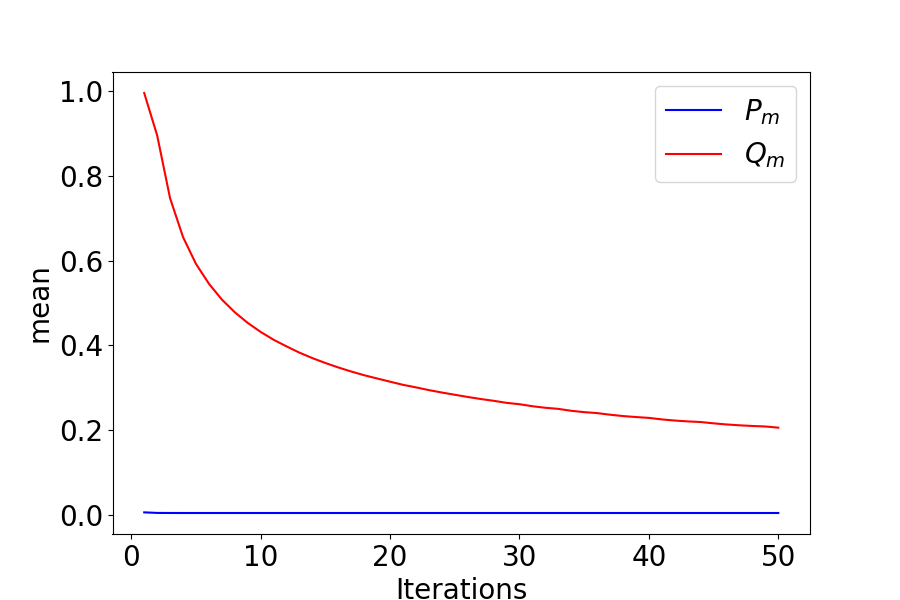}
    }
    \hfill
    \subfloat[The standard deviation.]{
        \includegraphics[width = 0.32\linewidth]{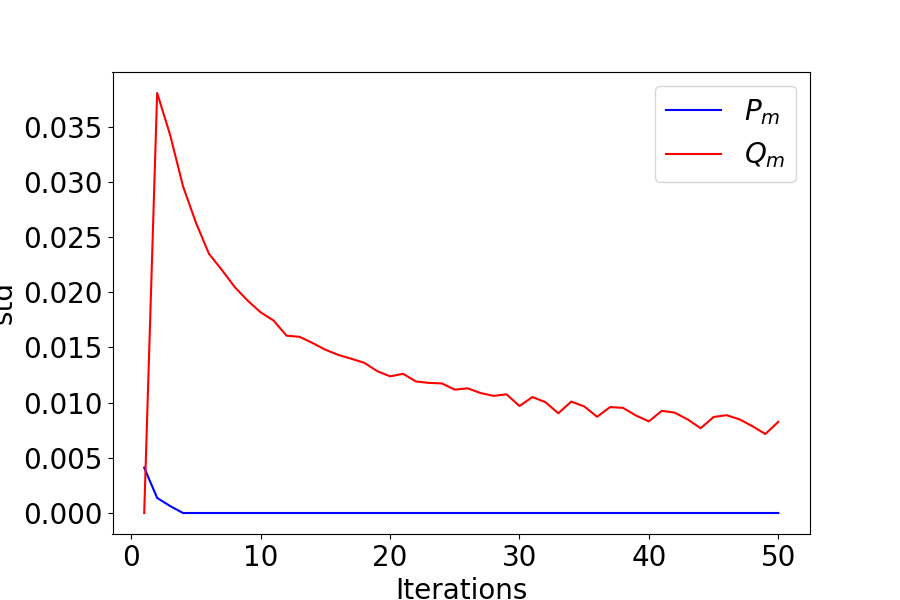}
    }
    \hfill
    \subfloat[The Wasserstein distance.]{
        \includegraphics[width = 0.32\linewidth]{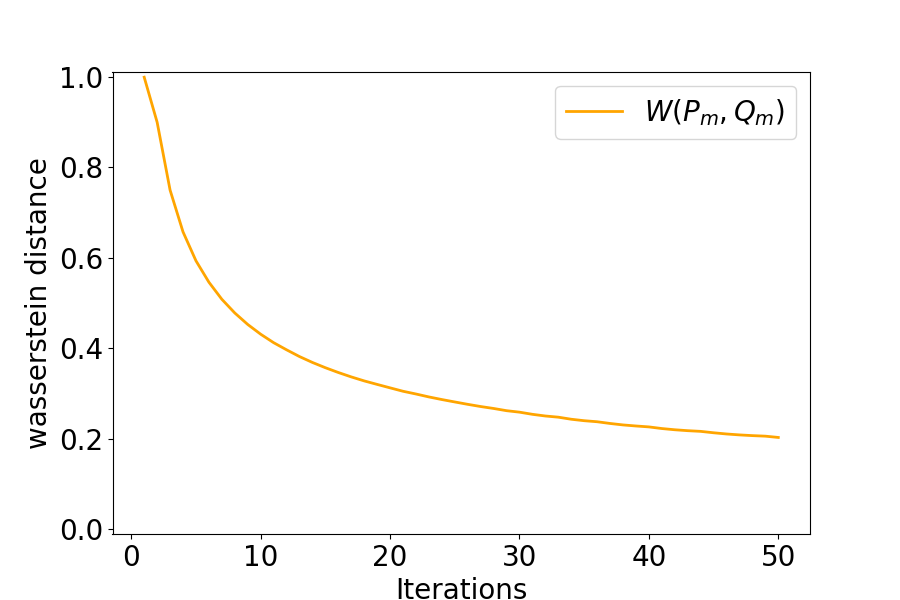}
    }
\caption{Quantitative analysis of the threshold function learning problem: (a) mean, (b) standard deviation, and (c) Wasserstein distance. }
\label{fig:TH_QA}
\end{figure*}

\begin{figure*}[ht]
    \centering
    \subfloat{
        \includegraphics[width = 0.32\textwidth]{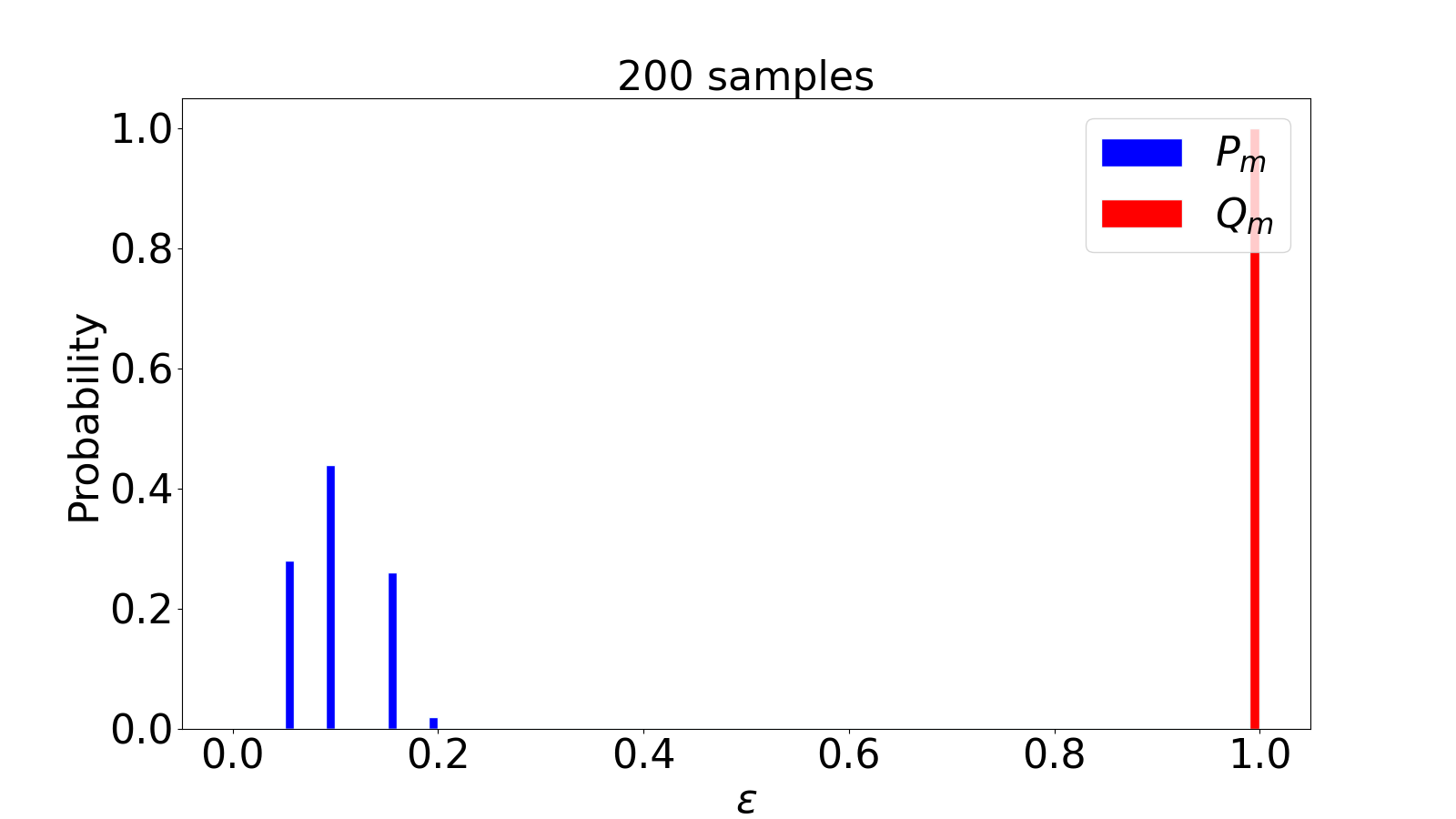}
    }
    \hfill
    \subfloat{
        \includegraphics[width = 0.32\textwidth]{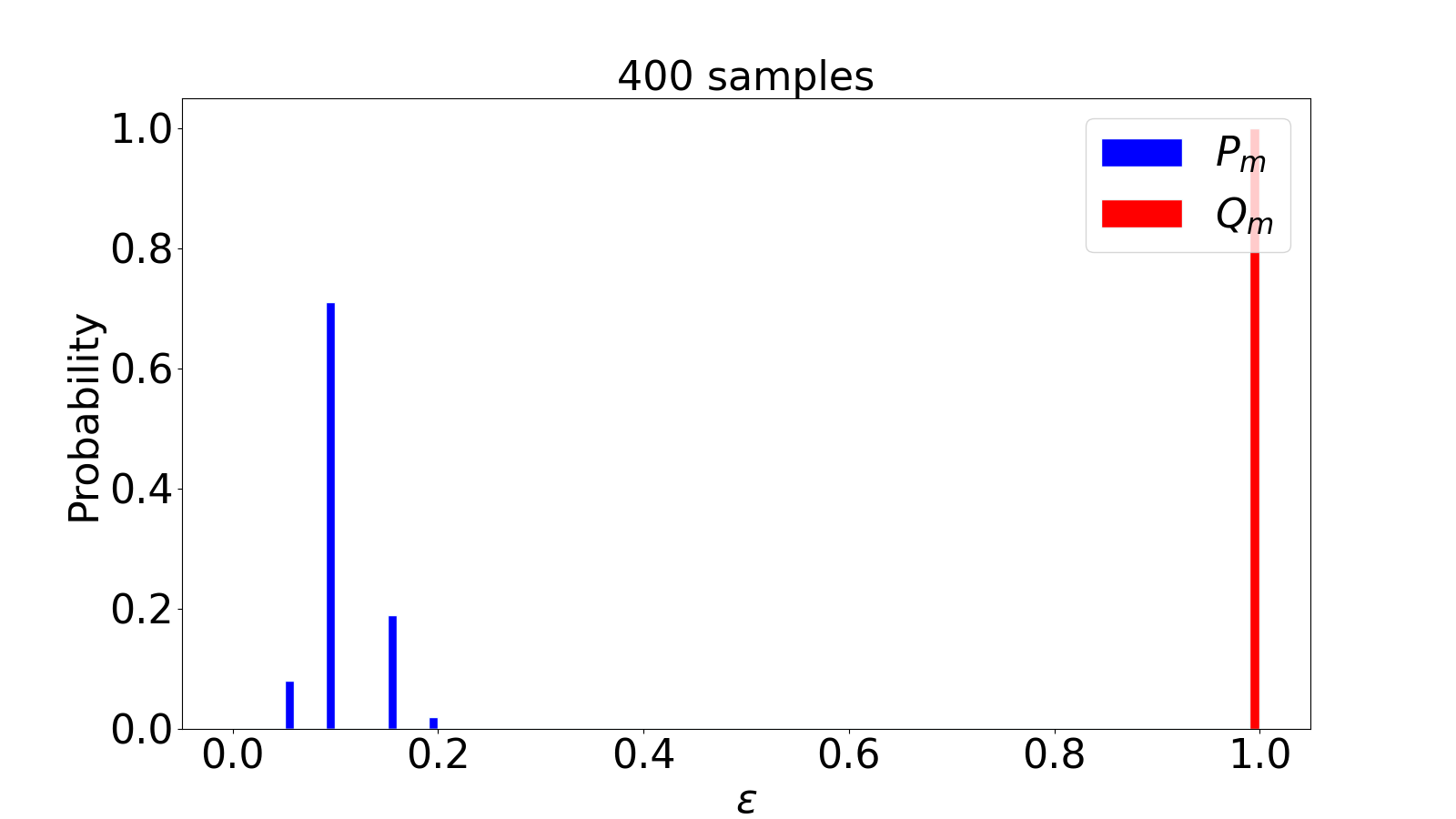}
    }
    \hfill
    \subfloat{
        \includegraphics[width = 0.32\textwidth]{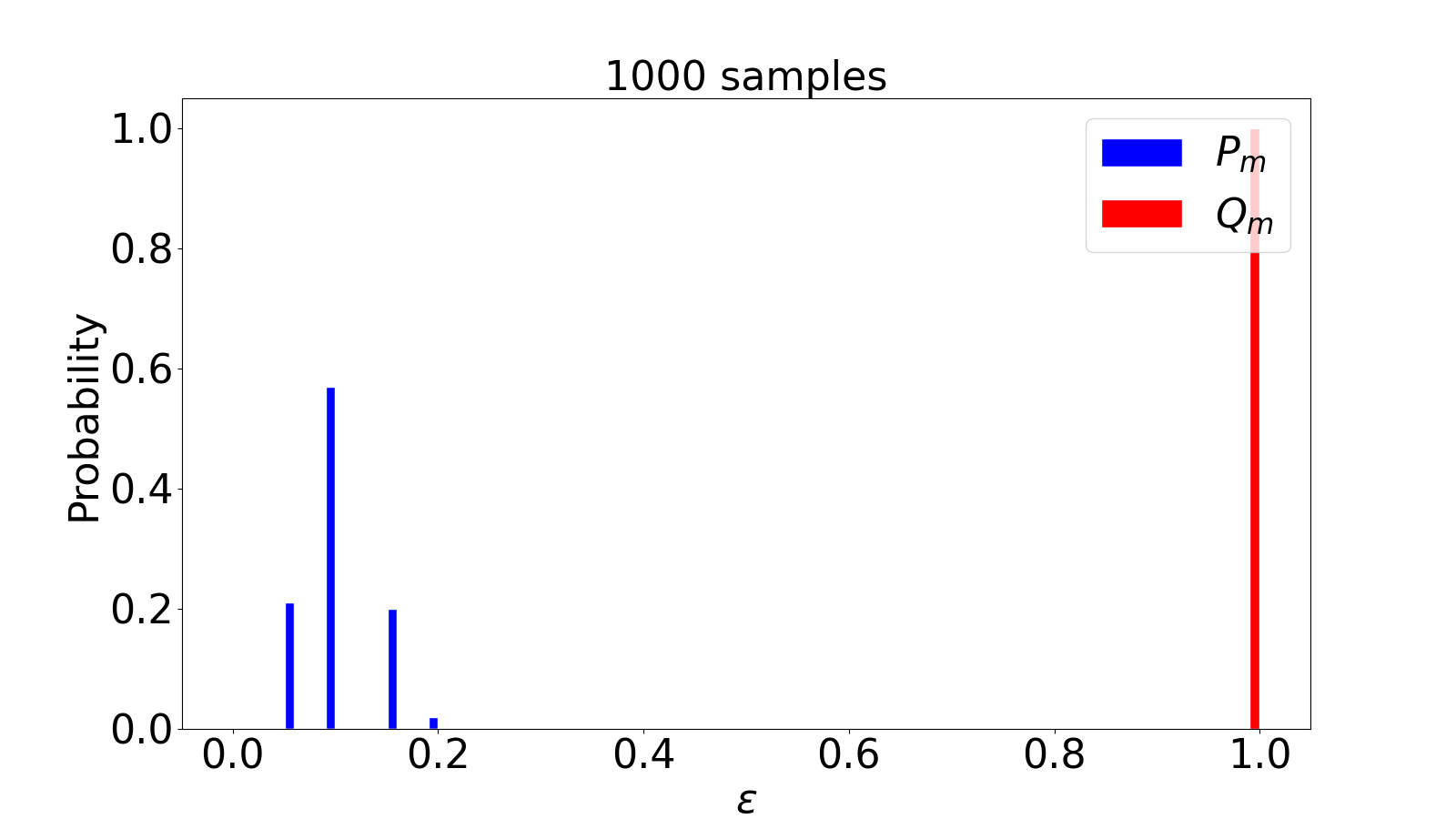}
    }
    \\
    \subfloat{
        \includegraphics[width = 0.32\textwidth]{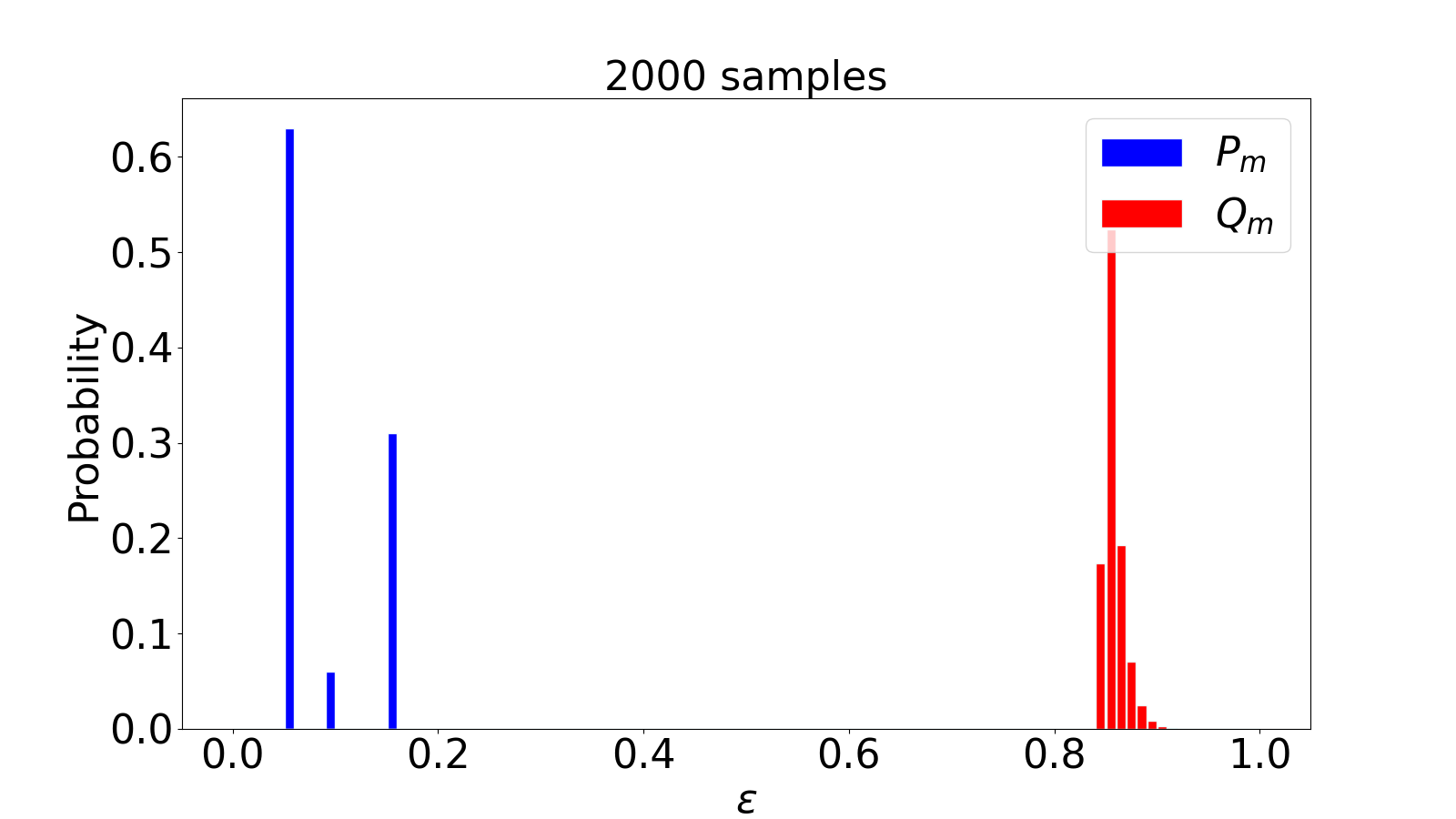}
    }
    \hfill
    \subfloat{
        \includegraphics[width = 0.32\textwidth]{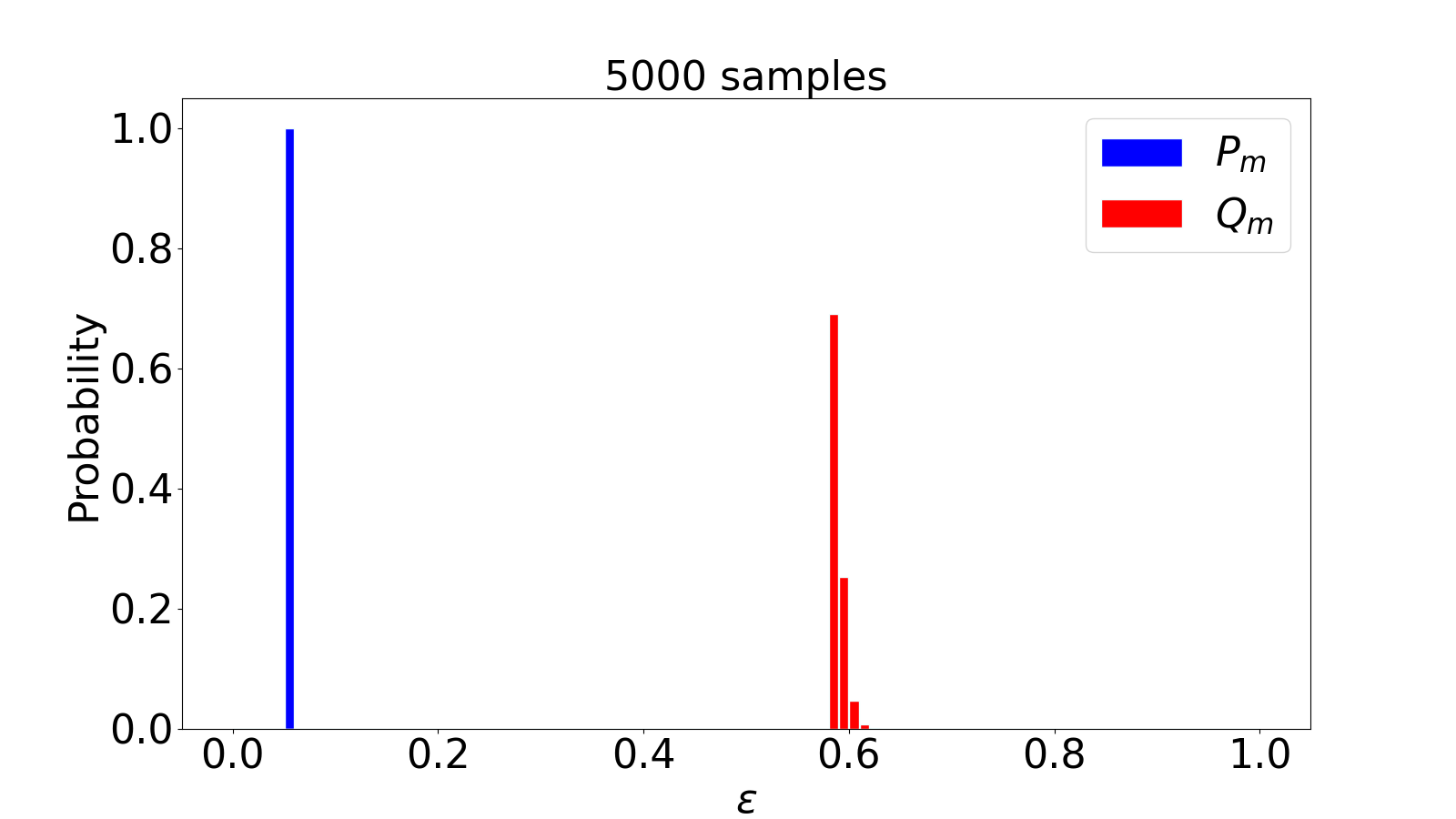}
    }
    \hfill
    \subfloat{
        \includegraphics[width = 0.32\textwidth]{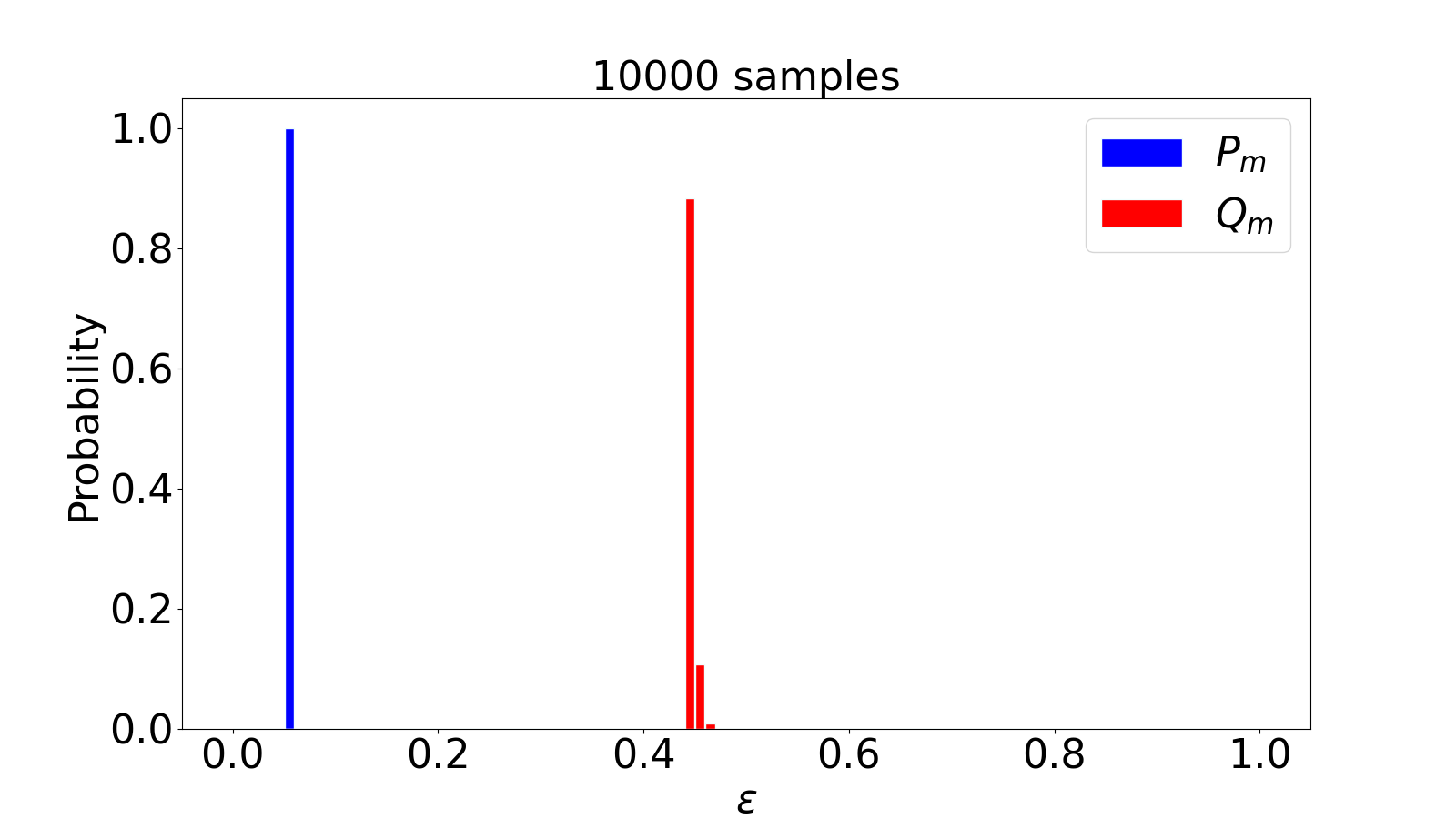}
    }
\caption{Distributions on the Iris data set with different sample size $m$.}
\label{fig:Iris}
\end{figure*}

To analyze $P_m$ and $Q_m$ with respect to sample size $m$, we ran the algorithm $50$ epochs, increasing the sample size $m$ at each round. For \cref{example:1}, the sample size is defined as $m = 25 \times iterations$. While for Example~\ref{example:2} and \ref{example:3}, $m$ is $200$ times the number of iterations.
For both tasks, we measure the time that is required to complete training on a single batch data set, 
and the overall runtime of the framework depends on the number of repeated samples
in each iteration and the total number of instances in each training sample. 
For the Boolean literal conjunction problem, the entire experiment was completed in approximately $121$ seconds, the threshold function problem required around $75$ seconds in total, and the Iris classification problem required around $107$ seconds in total. The empirical distributions $P_m$ (in \textcolor{blue}{blue}) and theoretical distributions $Q_m$ (in \textcolor{red}{red}) are shown in \cref{fig:CBL} 
for \cref{example:1} (conjunction of Boolean literals), 
\cref{fig:TH} for \cref{example:2} (threshold function),
and \cref{fig:Iris} 
for \cref{example:3} (Iris classification)~\footnote{For an agnostic PAC learnable problem, $\mu$ must be determined separately. In this example, we trained an SVM on the entire data set and took its test error as the value of $\mu$.}, respectively.

\begin{figure*}[t]
    \centering
    \subfloat[The mean.]{
        \includegraphics[width = 0.32\linewidth]{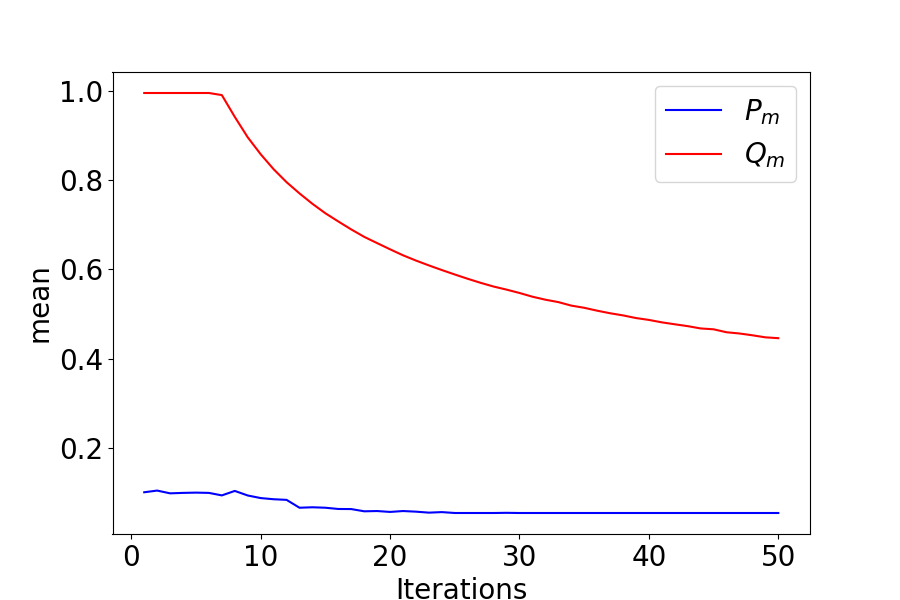}
    }
    \hfill
    \subfloat[The standard deviation.]{
        \includegraphics[width = 0.32\linewidth]{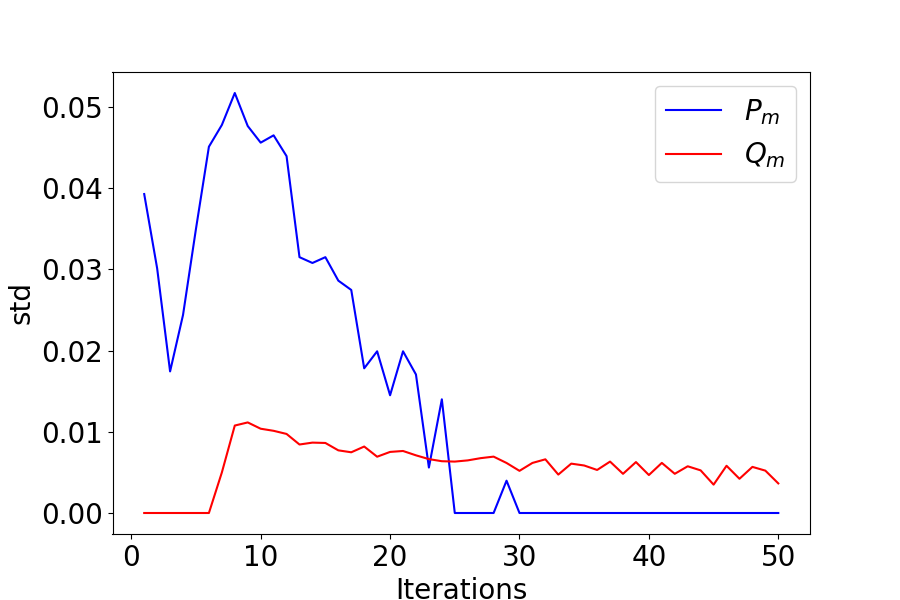}
    }
    \hfill
    \subfloat[The Wasserstein distance.]{
        \includegraphics[width = 0.32\linewidth]{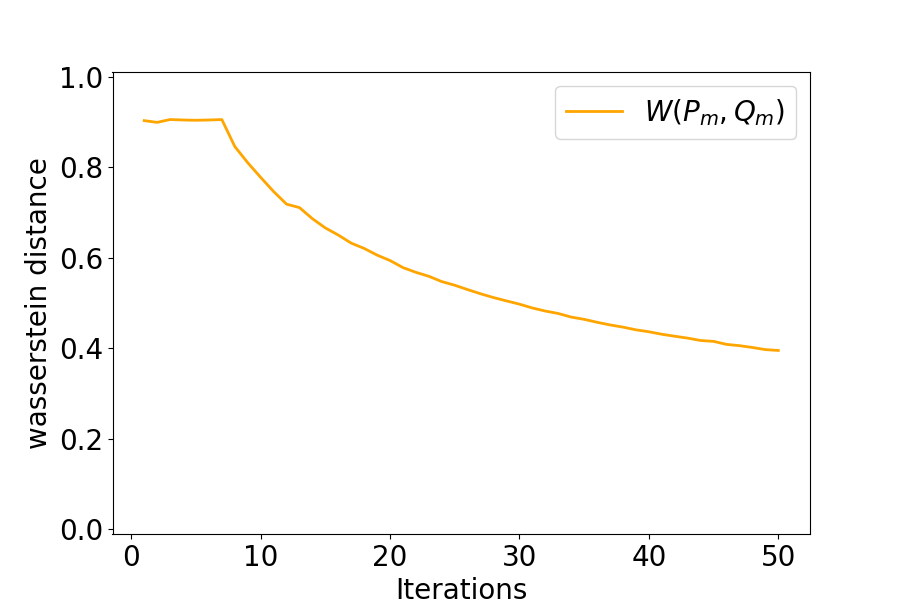}
    }
\caption{Quantitative analysis of Iris classification problem: (a) mean, (b) standard deviation, and (c) Wasserstein distance. 
}
\label{fig:Iris_QA}
\end{figure*}

\paragraph{Analysis on the results.}
Based on different sample sizes, we directly compare distributions $P_m$ and $Q_m$, as shown in \cref{fig:CBL}. One can easily observe that as the sample size increases, the distributions $P_m$ and $Q_m$ both shift towards left, which is consistent with our intuition. For all learning examples, the theoretical distributions $Q_m$ consistently remains to the right of $P_m$, before the two distributions start to merge for sufficiently large $m$. This result addresses RQ2 by showing that $Q_m$ serves as a conservative estimate for $P_m$. 

The learning curve for the Boolean literal conjunction learning problem is plotted in \cref{fig:CBL_QA} (a) - (c). 
In \cref{fig:CBL_QA}~(a), both the curves of $P_m$  and $Q_m$ exhibit a monotonic decrease, showing few upward fluctuations throughout the whole process. 
By quantitatively comparing losses between successive iterations, the mean of $P_m$ exhibited a monotonically decreasing trend with a probability of $96\%$ over $50$ iterations. 
This consistent downward trend indicates that the performance of both distributions improves steadily as the number of sample increases. 
Therefore, the answer to RQ1 and RQ2 are both positive for this PAC learnable problem. Additionally, \cref{fig:CBL_QA}~(b) illustrates that the standard deviation of the distributions $P_m$ and $Q_m$ fluctuates within a narrow range throughout the iterations. We apply Wasserstein distance as a measure to quantify how an empirical error distribution $P_m$ differs from the theoretical error distribution $Q_m$~\cite{arjovsky2017wasserstein}. 
In \cref{fig:CBL_QA}~(c), it shows a downtrend of Wasserstein distance as sample size increases.

For the threshold function learning problem, in a similar setting, we generate the data shown in \cref{fig:TH,fig:TH_QA}, with the only difference being that the theoretical error distributions are derived from \cref{eq:CDF-VC}, and the sample size is increased by $200$ in each round. 
Similar results for the Iris classification problem appear in \cref{fig:Iris,fig:Iris_QA}. 

\paragraph{Further discussions}

\begin{figure*}[!ht]
    \centering
    \subfloat[The mean.]{
        \includegraphics[width = 0.32\linewidth]{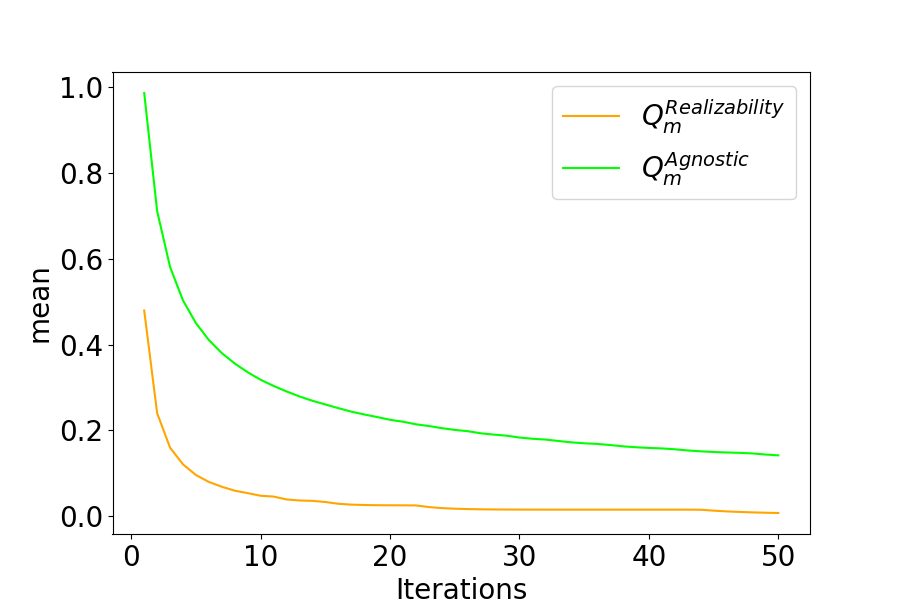}
    }
    \hfill
    \subfloat[The standard deviation.]{
        \includegraphics[width = 0.32\linewidth]{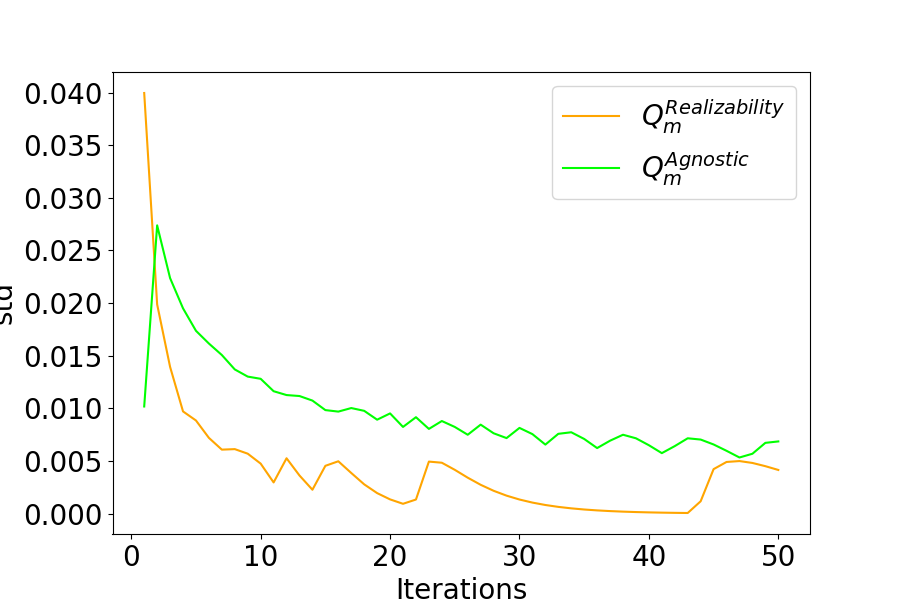}
    }
    \hfill
    \subfloat[The Wasserstein distance.]{
        \includegraphics[width = 0.32\linewidth]{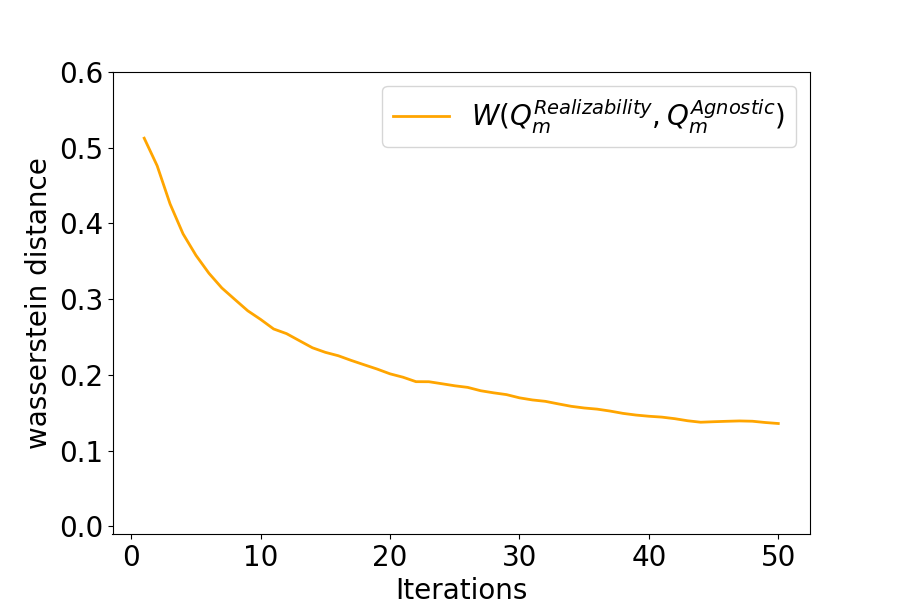}
    }
\caption{Quantitative analysis of the Boolean literal conjunction learning problem: (a) mean, (b) standard deviation, and (c) Wasserstein distance. 
Note that $Q_m^{Realizability}$, $Q_m^{Agnostic}$ are respectively based on \cref{eq:CDF} and \cref{eq:CDF-FHR}.}
\label{fig:CBL_QATH}
\end{figure*}

As discussed in \cref{sec:finite-H}, 
the set of realizably PAC learnable problems is a proper subset of agnostic PAC learnable problems.
While \cref{eq:CDF} provides a universal bound under the agnostic PAC framework, hypothesis set satisfying the realizability condition -- such as the case of Boolean literal conjunction problem -- may admit tighter sample complexity bounds through \cref{eq:CDF-FHR}, as demonstrated in \cref{fig:CBL_QATH}. 

Notably, similar reductions in sample complexity may also apply to hypothesis classes with finite VC dimension, though a systematic characterization of tight bounds under the realizability assumption remains an open problem. 
Crucially, Shalev-Shwartz and Ben-David also highlight that SVM algorithms achieve improved generalization bounds through geometric margin analysis, specifically by maximizing the margin-to-radius ratio in feature space (Section 15.2 in \cite{shalev2014understanding}). This observation underscores potential refinements in theoretical error distribution modeling.

\section{Related work}\label{sec:relatedwork}

Non-monotonic behavior in learning algorithms has been discussed in recent years~\cite{loog2019minimizers,viering2019open}, including several distinct phenomena. 
The earliest records for \emph{peaking} originate from pseudo-Fisher’s linear discriminant~\cite{vallet1989linear}, in which the authors find that a peak can appear when the sample size is close to the input dimension. 
\citet{belkin2019reconciling} provide further insights into this behavior, and their work suggests that the sample size at which the peak occurs represents an interpolation threshold between under-parameterization and over-parameterization. 
This research has gained some renewed attention under the name \emph{double descent}~\cite{belkin2019reconciling}. 
\emph{Dipping} has been observed in various tasks such as fitting binary feature data sets using the C4.5 algorithm~\cite{frey1999modeling}, outlier detection with $k$-nearest neighbor~\cite{ting2017defying}, and active learning for classification~\cite{konyushkova2015introducing}.
An explanation for this behavior is provided by \citet{chen2022explaining}, who analyze it through bias-variance decomposition. The third phenomenon, known as \emph{oscillatory behavior}, involves periodic fluctuations in generalization error, as discussed by \citet{loog2019minimizers}. 

In order to guarantee monotone learning, \citet{viering2020making} first propose a wrapper algorithm that compares the performance of a model trained with more data and the previously best model. 
Based on specific criteria, the algorithm determines whether to update the result, ensuring that the average generalization error obtained through multiple runs exhibits monotonicity with high probability. 
\citet{mhammedi2021risk} assumes that learning takes place in a more general statistical learning setting. 
By deriving new concentration inequalities, he demonstrates that an algorithm satisfying weak assumptions would ensure risk monotonicity, which holds for certain non-i.i.d. processes. 
\citet{bousquet2022monotone} introduce a novel monotonic algorithm aimed at transforming most learning algorithms into ones with similar performance. 
This transformation is achieved through a regularization process applied to both the current best model and the candidate model, where the current model is updated only if the candidate model exhibits a smaller error. 
While their method adopts a wrapper-like approach, its foundational principles are deeply influenced by the work of \citet{pestov2022universally}, which resolves a conjecture proposed in Devroye et al.'s book (Sect. 6.8)~\cite{devroye2013probabilistic} by providing a universally consistent learning rule that guarantees a monotonic decrease in the generalization error as the sample size increases. 
More recently, \citet{li2023monotonic} enhanced the wrapper algorithm by incorporating genetic algorithms and statistical methods, demonstrating the effectiveness of their approach on several popular data sets. 
The aforementioned methods primarily rely on wrapper algorithms and assume an inclusion relationship between neighboring data sets. Additionally, their monotonicity analyses are typically limited to specific statistics, such as the expected generalization error. In contrast, the sampling approach used in this paper is based on i.i.d. draws and imposes no such restrictions. Instead, we take advantage of the PAC framework which emphasizes conservative estimates of the generated hypotheses and naturally leads to monotonicity in the learning processes for PAC learnable tasks.

\section{Conclusion}\label{sec:conclusion}

This paper focused on the monotonicity of PAC learnable problems. We investigated two classes of machine learning problems: one with a finite hypothesis space and the other with a finite VC dimension. For both classes, we derived distributions of generalization error that serve as conservative theoretical bounds for evaluating the performance of generated models, demonstrating that these learning problems are theoretically monotone. %
To validate these findings, we conducted experiments on two simple PAC learnable problems. We compared the empirical error distributions ($P_m$) with their corresponding theoretical error distributions ($Q_m$) across an increasing series of sample sizes $m$. The experimental results confirmed two key conclusions for both learning problems: (1) the measured performance of generated models consistently improves as the sample size grows, and (2) the theoretical error distributions provide conservative estimates for the empirical error distributions of the generated models. Furthermore, as the sample size $m$ becomes sufficiently large, the empirical and theoretical error distributions converge to the smallest possible error for the hypothesis space $\Hs$.

\bibliographystyle{unsrtnat}
\bibliography{refs}

\newpage

\onecolumn


\appendix

\section{A derivation for the probability density function in Section~\ref{sec:finite-H}}
\label{sec:FHR-D}

We can derive the probability density function $f_{m}^{FHR}(\epsilon)$ corresponding to $F_{m}^{FHR}(\epsilon)$, wherein 
\begin{equation}
    F_{m}^{FHR}(\epsilon) = \int_{-\infty}^{\epsilon} f_{m}^{FHR}(x) dx.
\label{eq:integral-FHR}
\end{equation}
However, it should be noted that $f$ exhibits discontinuity at $\epsilon$, thus we need to use the Dirac delta function $\delta(x)$ to obtain $f_{m}^{FHR}(x)$. 
Regarding $\delta(x)$, there exist two equivalent definitions as follows: 
\begin{equation}
\begin{matrix}
    \delta(x) = 0, x \neq 0, & \int_{-\infty}^{+\infty} \delta(x) dx = 1;
\end{matrix}
\label{eq:delta-1}
\end{equation}
and
\begin{equation}
\begin{matrix}
    \delta(x) = \frac{dH(x)}{dx}, & H(x) = \begin{cases} 0, & x < 0 \\ 1, & x \geq 0 \end{cases}.
\end{matrix}
\label{eq:delta-2}
\end{equation}

Consequently, we can decompose $F_{m}^{FHR}(\epsilon)$ into continuous and discontinuous components, and separately establish
\begin{equation}
    H_{m}^{FHR}(\epsilon) \overset{\text{def}}{=} 
    \begin{cases}
    0 & \epsilon < 1 \\
    |\Hs| \exp{(- m)} & \epsilon \geq 1 \\
    \end{cases}.
\end{equation}
and 
\begin{equation}
    G_{m}^{FHR}(\epsilon) \overset{\text{def}}{=} 
    \begin{cases}
    0 & \epsilon < \frac{\ln(|\Hs|)}{m} \\
    1 - |\Hs| \exp{(- m \epsilon)} & \frac{\ln(|\Hs|)}{m} \leq \epsilon < 1 \\
    1 - |\Hs| \exp{(- m)} & \epsilon \geq 1 \\
    \end{cases}.
\end{equation}
Then $F_{m}^{FHR}(\epsilon) = G_{m}^{FHR}(\epsilon) + H_{m}^{FHR}(\epsilon)$, where $G_{m}^{FHR}(\epsilon)$ is continuous and $H_{m}^{FHR}(\epsilon)$ is a step function. 
Therefore, according to \cref{eq:integral-FHR} we can obtain a probability density function corresponding to the probability distribution function $F_{m}(\epsilon)$, such as
\begin{equation}
\begin{split}
    f_{m}^{FHR}(\epsilon) & = \frac{dF_{m}^{FHR}(\epsilon)}{d\epsilon} \\
    & = \frac{dG_{m}^{FHR}(\epsilon)}{d\epsilon} + \frac{dH_{m}^{FHR}(\epsilon)}{d\epsilon} \\
    & = \frac{dG_{m}^{FHR}(\epsilon)}{d\epsilon} + |\Hs| \exp{(- m)} \frac{dH(\epsilon - 1)}{d\epsilon} \\
    & = \frac{dG_{m}^{FHR}(\epsilon)}{d\epsilon} + |\Hs| \exp{(- m)} \delta(\epsilon - 1)
\end{split}
\end{equation}
where $\frac{dF_{m}^{FHR}(\epsilon)}{d\epsilon}$ and $\frac{dG_{m}^{FHR}(\epsilon)}{d\epsilon}$ respectively refer to a weak derivative of $F_{m}^{FHR}$ and $G_{m}^{FHR}$, and
\begin{equation}
    \frac{dG_{m}^{FHR}(\epsilon)}{d\epsilon} = 
    \begin{cases}
    0 & \epsilon < \frac{\ln(|\Hs|)}{m} \; or \; \epsilon \geq 1 \\
    |\Hs| m \exp{(- m \epsilon)} & \frac{\ln(|\Hs|)}{m} \leq \epsilon < 1
    \end{cases}.
\end{equation}

\section{The sample complexity of PAC Learnability for a finite VC dimension.
}
\label{sec:SC-VCAD}

Given a hypothesis class, $\Hs$, the ERM learner selects a member of $\Hs$ that minimizes the empirical risk on the training sample $S$. 
We wish that this $h$ not only minimizes the empirical risk but also closely approximates the true data probability distribution $\D$. 
Therefore, we need that uniformly over all hypotheses in the hypothesis class, the empirical risk will be close to the true risk, as formalized in the following. 

\begin{Definition} ($\epsilon$-representative sample)~\cite{shalev2014understanding}
A training set $S$ is called $\epsilon$-representative (w.r.t. domain $\Z = \X \times \Y$, hypothesis class $\Hs$, loss function $\loss$, and distribution $\D$) if
\begin{equation}
    \forall h \in \Hs, |L_{S}(h) - L_{\D}(h)| \leq \epsilon.
\end{equation}
\end{Definition}

The above observation motivates the following lemma. 

\begin{Lemma} \cite{shalev2014understanding}
Assume that a training set $S$ is $\frac{\epsilon}{2}$-representative (w.r.t. domain $\Z$, hypothesis class $\Hs$, loss function $\loss$, and distribution $\D$). Then, any output of $ERM_{\Hs}(S)$, namely, any $h_S \in \arg\min_{h' \in \Hs} L_{S}(h)$, satisfies 
\begin{equation}
    L_{\D}(h_S) \leq \min_{h' \in \Hs} L_{D}(h') + \epsilon.
\end{equation}
\label{lem:UC}
\end{Lemma}

From this lemma, it follows that ERM will be an agnostic PAC learner if one can demonstrate that, with probability at least $1 - \delta$ over the random choice of the training set, the set is $\epsilon / 2$-representative. The uniform convergence condition provides the formal statement of this requirement. 

\begin{Definition} (Uniform Convergence)~\cite{shalev2014understanding} 
We say that a hypothesis class $\Hs$ has the uniform convergence property (w.r.t. a domain $\Z$ and a loss function $\loss$) if there exists a function $m_{\Hs}^{UC}: {(0, 1)}^2 \rightarrow \Nat$ such that for every $\epsilon, \delta \in (0, 1)$ and for every probability distribution D over $\Z$, if $S$ is a sample of $m \geq m_{\Hs}^{UC}(\epsilon, \delta)$ examples drawn i.i.d. according to $\D$, then, with probability of at least $1 - \delta$, $S$ is $\epsilon$-representative.
\end{Definition}

To show that uniform convergence holds, we first introduce \cref{lem:GF} that relates the growth function to the generalization error. 
\begin{Lemma} 
    For a hypothetical space $\Hs$, $h \in \Hs$, $m \in \Nat$, and $0 < \epsilon < 1$. When $m > 2/\epsilon^2$, we have that
    \begin{equation}
        \P(|L_{\D}(h) - L^{*}| > \epsilon) \leq 4 \Pi_{\Hs}(2m) \exp{(- \frac{m \epsilon^2}{8})},
    \end{equation}
\label{lem:GF}
\end{Lemma}

According to Sauer’s lemma~\cite{sauer1972density} and \cref{lem:UC}, we can obtain \cref{cor:CI-VC}.

\section{A derivation for the probability density function in Section~\ref{sec:finite-VC}}
\label{sec:VCAD}



By the definition of the probability density function $f_{m}^{VC}(x)$, we have 
\begin{equation}
    F_{m}^{VC}(\epsilon) = \int_{-\infty}^{\epsilon} f_{m}^{VC}(x) dx.
\label{eq:integral_VC}
\end{equation}

Because the function is discontinuous, we must introduce the the Dirac delta function $\delta(x)$ here; it was defined in \cref{eq:delta-1,eq:delta-2}. 
Next, we can decompose $F_{m}^{VC}(\epsilon)$ into continuous and discontinuous components, and separately establish
\begin{equation}
    H_{m}^{VC}(\epsilon) \overset{\text{def}}{=} 
    \begin{cases}
    0 & \epsilon < 1 - \mu \\
    4 (\frac{2em}{d})^d \exp{(- \frac{m (1 - \mu)^2}{32})} & \epsilon \geq 1 - \mu \\
    \end{cases},
\end{equation}
and 
\begin{equation}
    G_{m}^{VC}(\epsilon) \overset{\text{def}}{=} 
    \begin{cases}
    0 & \epsilon < \sqrt{\frac{32 d \ln{(2em/d)} + 32\ln{(4)}}{m}} \\
    1 - 4 (\frac{2em}{d})^d \exp{(- \frac{m \epsilon^2}{32})} & \sqrt{\frac{32 d \ln{(2em/d)} + 32\ln{(4)}}{m}} \leq \epsilon < 1 - \mu \\
    1 - 4 (\frac{2em}{d})^d \exp{(- \frac{m (1 - \mu)^2}{32})} & \epsilon \geq 1 - \mu \\
    \end{cases}.
\end{equation}
It follows that $F_{m}^{VC}(\epsilon) = G_{m}^{VC}(\epsilon) + H_{m}^{VC}(\epsilon)$, where $G_{m}(\epsilon)$ is continuous and $H_{m}(\epsilon)$ is a step function. 
Therefore, according to \cref{eq:integral_VC} we can obtain a probability density function corresponding to the probability distribution function $F_{m}(\epsilon)$, such as
\begin{equation}\label{eq:PDF-FVC}
\begin{split}
    f_{m}^{VC}(\epsilon) & = \frac{dF_{m}^{VC}(\epsilon)}{d\epsilon} \\
    & = \frac{dG_{m}^{VC}(\epsilon)}{d\epsilon} + \frac{dH_{m}^{VC}(\epsilon)}{d\epsilon} \\
    & = \frac{dG_{m}^{VC}(\epsilon)}{d\epsilon} + 4 (\frac{2em}{d})^d \exp{(- \frac{m (1 - \mu)^2}{32})} \frac{dH_{m}^{VC}(\epsilon - (1 - \mu))}{d\epsilon} \\
    & = \frac{dG_{m}^{VC}(\epsilon)}{d\epsilon} + 4 (\frac{2em}{d})^d \exp{(- \frac{m (1 - \mu)^2}{32})} \delta(\epsilon - (1 - \mu))
\end{split}
\end{equation}
where $\frac{dF_{m}^{VC}(\epsilon)}{d\epsilon}$ and $\frac{dG_{m}^{VC}(\epsilon)}{d\epsilon}$ respectively refer to a weak derivative of $F_{m}$ and $G_{m}^{VC}$, and
\begin{equation}
    \frac{dG_{m}^{VC}(\epsilon)}{d\epsilon} = 
    \begin{cases}
    0 & \epsilon < \frac{d}{m} \sqrt{\frac{32 d \ln{(2em/d)} + 32\ln{(4)}}{m}}  \; or \; \epsilon \geq 1 - \mu \\
    4 (\frac{2em}{d})^d \exp{(- \frac{m (1 - \mu)^2}{32})} & \sqrt{\frac{32 d \ln{(2em/d)} + 32\ln{(4)}}{m}} \leq \epsilon < 1 - \mu
    \end{cases}.
\end{equation}

\end{document}